%% file: main.tex
\documentclass[10pt,twocolumn,letterpaper]{article}

\usepackage{iccv}              

\usepackage{overpic}
\usepackage{colortbl}
\usepackage{multirow}

\definecolor{grayblue}{rgb}{0.184, 0.332, 0.589}
\definecolor{yellow}{rgb}{0.746, 0.563, 0.0}
\definecolor{green}{rgb}{0.402, 0.566, 0.25}
\newcommand{\gr}{\rowcolor[gray]{.95}}


\definecolor{iccvblue}{rgb}{0.21,0.49,0.74}
\usepackage[pagebackref,breaklinks,colorlinks,allcolors=iccvblue]{hyperref}
\usepackage{marvosym}

\def\paperID{*****} 
\def\confName{ICCV}
\def\confYear{2025}

\title{ACMamba: Fast Unsupervised Anomaly Detection via An Asymmetrical Consensus State Space Model}

\author{Guanchun Wang$^1$, Xiangrong Zhang$^1$$^{(\textrm{\Letter})}$, Yifei Zhang$^1$, Zelin Peng$^2$, Tianyang Zhang$^1$, \and Xu Tang$^1$, Licheng Jiao$^1$\\
$^1$School of Artificial Intelligence, Xidian University\\
$^2$MoE Key Lab of Artificial Intelligence, AI Institute, Shanghai Jiao Tong University\\
\small{\tt{wangguanchun@xidian.edu.cn, xrzhang@mail.xidian.edu.cn}} \\ 
}

\begin{document}
\maketitle
\input{sec/0_abstract}    
\input{sec/1_intro}
\input{sec/2_formatting}
{
    \small
    \bibliographystyle{ieeenat_fullname}
    \bibliography{main}
}

\input{sec/X_suppl}

\end{document}

%% file: sec/0_abstract.tex
\begin{abstract}
Unsupervised anomaly detection in hyperspectral images (HSI), aiming to detect unknown targets from backgrounds, is challenging for earth surface monitoring. However, current studies are hindered by steep computational costs due to the high-dimensional property of HSI and dense sampling-based training paradigm, constraining their rapid deployment. Our key observation is that, during training, not all samples within the same homogeneous area are indispensable, whereas ingenious sampling can provide a powerful substitute for reducing costs. Motivated by this, we propose an Asymmetrical Consensus State Space Model (ACMamba) to significantly reduce computational costs without compromising accuracy. Specifically, we design an asymmetrical anomaly detection paradigm that utilizes region-level instances as an efficient alternative to dense pixel-level samples. In this paradigm, a low-cost Mamba-based module is introduced to discover global contextual attributes of regions that are essential for HSI reconstruction. Additionally, we develop a consensus learning strategy from the optimization perspective to simultaneously facilitate background reconstruction and anomaly compression, further alleviating the negative impact of anomaly reconstruction. Theoretical analysis and extensive experiments across eight benchmarks verify the superiority of ACMamba, demonstrating a faster speed and stronger performance over the state-of-the-art.
\end{abstract}

%% file: sec/1_intro.tex
\section{Introduction}
\label{sec:intro}

Anomaly detection in earth surface monitoring, aiming to detect unknown targets that differ from normal backgrounds without supervision, has recently gained prominence in precision agriculture, disaster early warning, and ecological environmental monitoring \cite{HAD-review}. Hyperspectral image (HSI), a high-dimensional data consisting of substantial spectral bands offering intricate details of land cover materials \cite{NIPS-HSI,AAAI-HSI}, are ideal for detecting anomaly targets, thereby hyperspectral anomaly detection (HAD) has emerged as a significant branch of anomaly monitoring.

\begin{figure}[t]
    \centering
    \begin{overpic}[width=\linewidth]{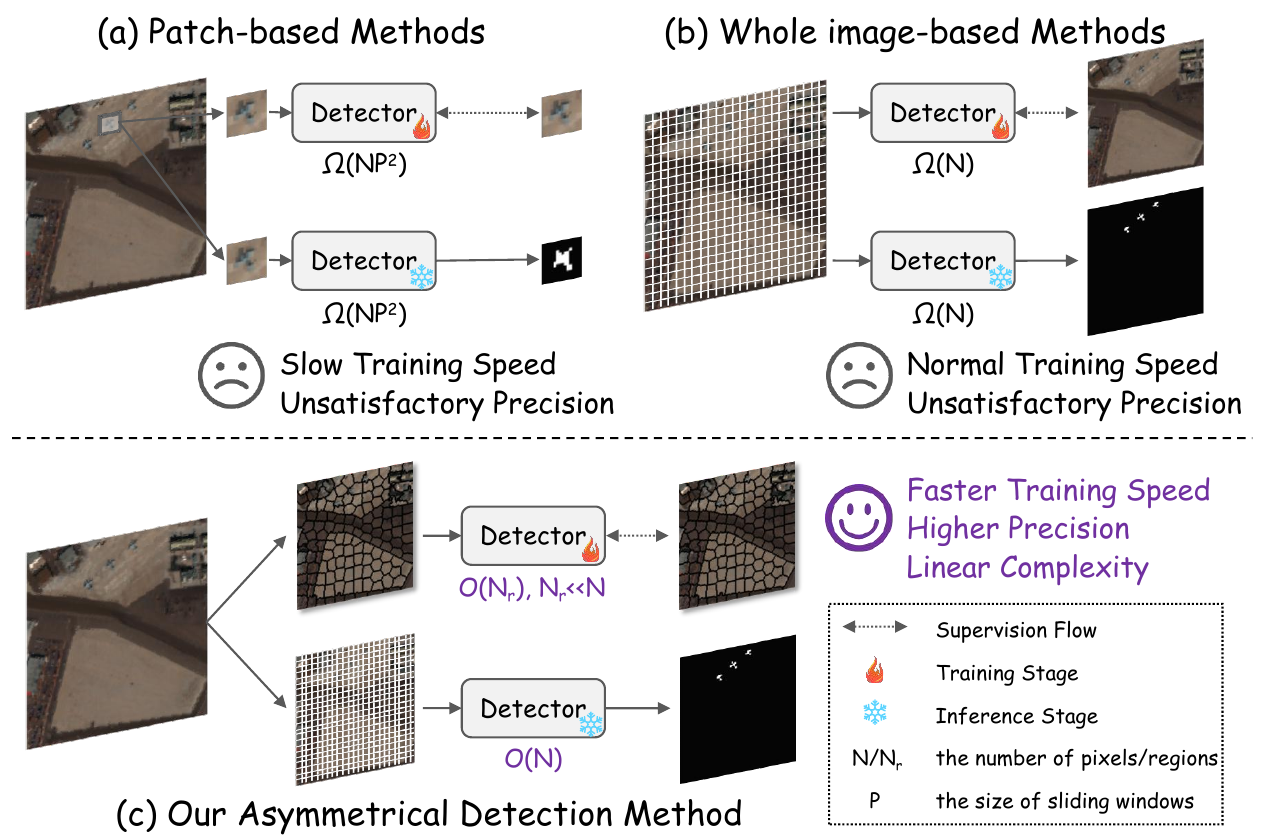}
    \end{overpic}
    \caption{Comparison between previous methods (top) and ACMamba (bottom). Patch- and whole image-based methods are restricted with heavy costs due to dense sampling, \emph{i.e.,} $\Omega(N P^2)$ and $\Omega(N)$. Our ACMamba mitigates it by an asymmetrical detection paradigm, relieving computations to $\Omega(N_\text{r})$, $N_\text{r} \ll N$.}
    \label{fig:intro}
\end{figure}
Traditional methods typically separate the background and anomaly distributions by leveraging statistics-based \cite{RX,LRX} or representation-based \cite{LRCRD,PTA,PCA-TLRSR} models, achieving remarkable performance but struggling with manually setting hyperparameters across different scenarios. Thanks to the impressive representation capability of visual foundation models, deep learning-based methods \cite{LREN,GT-HAD} have been prevalently explored to automatically detect anomaly targets.

Depending on how to feed HSI data into the deep neural network, existing methods can be roughly divided into two categories, \emph{i.e.,} patch-based \cite{GT-HAD,DirectNet,AUD-Net} and whole image-based \cite{LREN,autoAD,DFAN-HAD} methods. The former primarily focuses on local detection by sliding windows, which brings a considerable computational cost due to the duplicated inference of overlapping pixels. Another deep learning-based method, employing the whole image as the input \cite{LREN,autoAD, DFAN-HAD}, directly learns a point-to-point semantic mapping but still involves extensive computations per epoch. As shown in Figure~\ref{fig:intro}, the patch-based and whole image-based detectors require $\Omega(NP^2)$ and $\Omega(N)$ calculations per epoch during training, respectively, which are frequently limited by computational efficiency.  Nevertheless, we observe that not all instances within a homogeneous region are essential, which is attributed to their analogous characteristic. \textit{\textbf{This motivates us to leverage sparse representative samples to intensely compress training costs.}}





Moreover, global contextual information is crucial for anomaly detection. Transformer havs emerged as a powerful selection for global modeling but are restricted to quadratic computational complexity \cite{transformer, ViT}. Low-cost foundational models have recently demonstrated their efficacy in multiple fields \cite{Mamba,RWKV,Mamba2}, specifically for the Mamba series \cite{Vmamba, Visionmamba} distinguished by selective state space models. \textit{\textbf{This inspires us to build a low-cost detector capable of meticulous HSI anomaly detection.}} Besides, prevalent methods predominantly detect targets through reconstruction error maps, where the crucial is to facilitate background reconstruction and hamper anomalous regions. Data masking strategies are widely employed to avoid memorizing anomalous targets, however, such methods only prevent the model from encountering anomalies while lacking specific guidance of anomaly targets, leading to a risk of missing detections due to spectral shifts. This naturally arises: \textbf{\textit{How to ensure the equivalent reconstruction of backgrounds while guiding anomalies into a collapsed space?}}

Considering the above issues, this paper proposes a fast \textit{\textbf{Asymmetrical Consensus State Space Model}} for hyperspectral anomaly detection, termed \textit{\textbf{ACMamba}}. Specifically, ACMamba is first constructed with an asymmetrical anomaly detection (AAD) paradigm, which solely requires a single representative sample in each homogeneous region for training but precisely detects dense pixels, significantly reducing the training cost and equipping spatial geometric cues. Under this paradigm, a regional spectral attribute learning (RSAL) module equipped with selective structure state space models is designed to capture global contextual information, thereby detecting anomaly regions in a low-cost manner. To alleviate the negative impact of anomaly reconstruction, we additionally introduce a consensus learning strategy (CLS), which projects anomalous targets into a collapsed space while ensuring background recovery. The contribution of our work can be summarized as follows:
    
\begin{itemize} 
\item We propose a fast hyperspectral anomaly detector dubbed ACMamba, which breaks the bottleneck of deployment costs by an asymmetric anomaly detection (AAD) paradigm, revealing that meticulous anomaly detection can be performed without dense sampling training.


\item We design a regional spectral attribute learning (RSAL) module with selective structure state space models to efficiently discover global contexts of HSIs.

\item We present a consensus learning strategy (CLS) to optimize ACMamba, ensuring backgrounds reconstruction while facilitating anomalies into a collapsed space.

\item Theoretical analysis and experimental results jointly demonstrate that our ACMamba can achieve state-of-the-art performance with the fastest detection speed.
\end{itemize}

%% file: sec/2_formatting.tex
\section{Related Work}
\subsection{Hyperspectral Anomaly Detection}
Hyperspectral anomaly detection emerged from the challenge of recognizing unknown targets that differ from surrounding scenes without labeled data, which can be divided into traditional and deep learning-based methods.

\subsubsection{Traditional Methods}
Traditional methods typically address it by statistics- \cite{RX,LRX,2S-GLRT,KIFD,MsRFQFT} or representation-based \cite{CRD,LRCRD,PTA,PCA-TLRSR} techniques. Reed et al. \cite{RX} introduced the Reed–Xiaoli (RX) method, a classical statistics-based method following the generalized likelihood ratio test, which assumes that backgrounds in HSI follow a multivariate Gaussian distribution and detects anomaly targets by measuring the Mahalanobis distance between pixels. Various improved methods leveraging RX as the basic pipeline have been proposed, such as local RX \cite{LRX}, weighted RX \cite{WRX}, subspace RX \cite{SRX}, and two-step generalized likelihood ratio test method (2S-GLRT) \cite{2S-GLRT}. Representation-based methods generally assume that each pixel in backgrounds can be roughly represented by its neighborhoods in space and consider pixels that cannot be adequately described as anomaly targets. Li et al. \cite{CRD} proposed a collaborative representation-based detector (CRD), which adopts a dual-window strategy that utilizes pixels in the outer window to detect the central pixel by linear combining reconstruction. Based on CRD, Su et al. \cite{LRCRD} considered the low-rank properties of backgrounds and the sparsity of anomaly and proposed a low-rank and collaborative representation detector (LRCRD), ensuring the accuracy and stability of anomaly detection. In addition, Li et al. \cite{PTA} and Wang et al. \cite{PCA-TLRSR} followed low-rank and sparse-based tensor approximation to better separate backgrounds and anomalies. \textit{Although achieving remarkable results, these methods require modifying hyperparameters in different scenarios, such as window size, leading to inconvenience in practical application.}
\begin{figure*}[t]
    \centering
    \begin{overpic}[width=\linewidth]{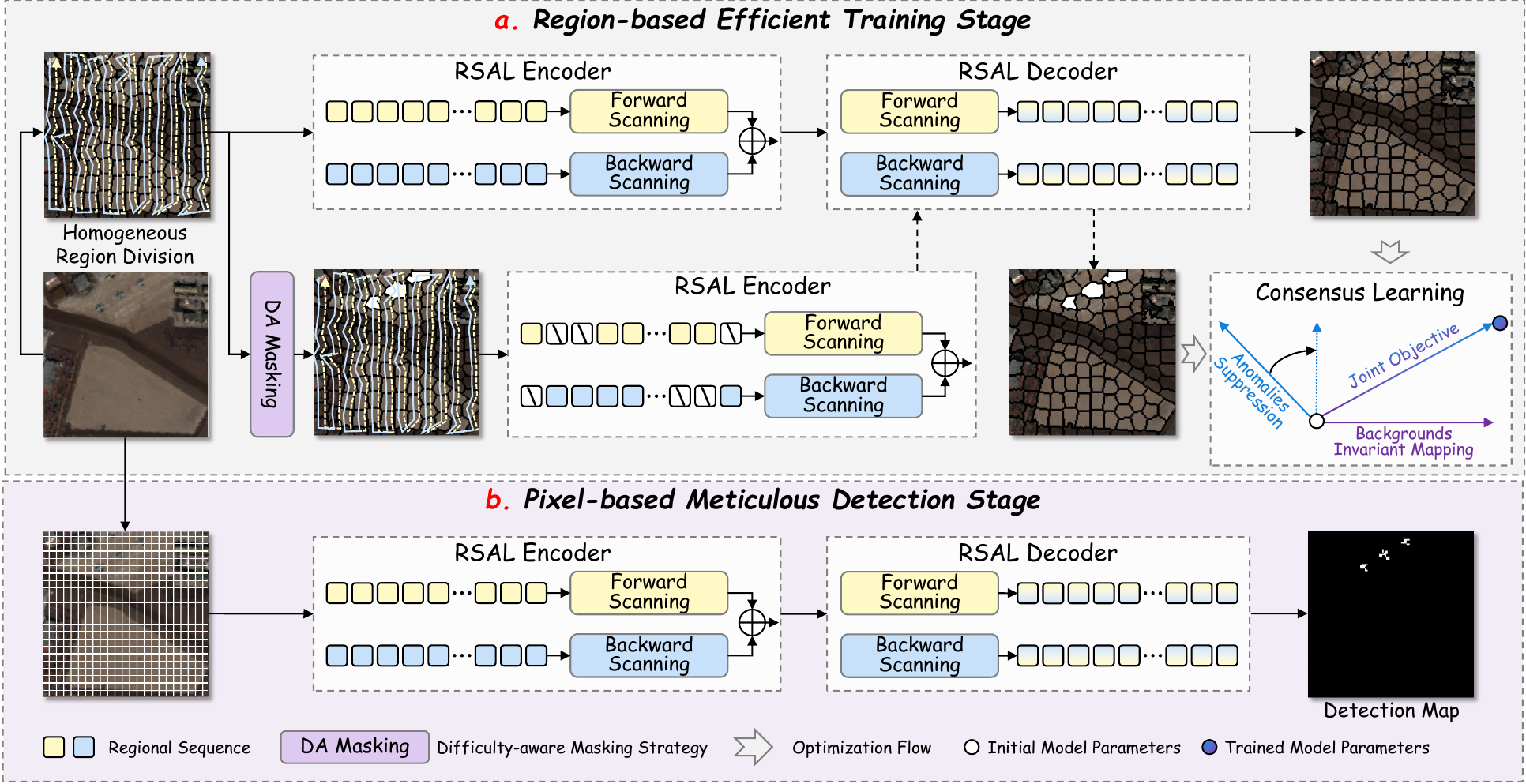}
    \put(88, 49){\scriptsize$\mathcal D({\mathcal E}(\mathbf {\tilde X}))$}
    \put(66.5, 21){\scriptsize$\mathcal D(\hat{\mathcal E}(\mathbf M\odot \mathbf {\tilde X}))$}
    \put(23.5, 21){\scriptsize$\mathbf M\odot \mathbf {\tilde X}$}
    \end{overpic}
    \caption{Illustration of ACMamba that adopts an asymmetric anomaly detection (AAD) paradigm including a region-based efficient training stage and a pixel-based meticulous detection stage. At each stage, a lightweight autoencoder incorporated with regional spectral attribute learning (RSAL) modules is utilized for HSI reconstruction. A consensus learning strategy (CLS) is used to facilitate background reconstruction and anomaly compression, further improving detection performance.}
    \label{fig:overall}
\end{figure*}

\subsubsection{Deep learning-based Methods}
With the recent success of deep learning techniques \cite{DL}, various studies have employed them for hyperspectral anomaly detection. One of the primary distinctions among these methods is how to feed HSIs into deep neural networks, which can be roughly divided into two categories, e.g., patch-based \cite{LS3TNet,GT-HAD,DirectNet,BS3LNet,PDBSNet,AUD-Net,TAEF} and whole image-based \cite{autoAD,RGAE,MSNet,DFAN-HAD,LREN,DeepLR,SSHAD} methods.

Patch-based methods \cite{LS3TNet, AUD-Net,BS3LNet,DirectNet,PDBSNet, GT-HAD, TAEF} often split the image into numerous overlapping regions and successively feed them into the DL model for feature extraction and detection, where the information within each patch is utilized to determine whether its center pixel is an anomaly. Since most previous methods lack extraction of local spatial information, Wang et al. \cite{LS3TNet} proposed a dual-stream detector based on local spatial-spectral information aggregation, which adopts adaptive convolutions and fully connected layers to mine spatial knowledge. To satisfactorily reconstruct pure backgrounds, Gao et al. \cite{BS3LNet} and Wang et al. \cite{DirectNet,PDBSNet} proposed a series of hyperspectral anomaly detectors based on blind-spot networks, which reduce the awareness of the center pixel in the HSI patch, thereby preventing the reconstruction of anomaly targets. Moreover, Lian et al. \cite{GT-HAD} proposed a gated Transformer detector called GT-HAD, which enhances background features and suppresses anomaly features by using spatial-spectral content similarity.

Whole image-based methods \cite{autoAD,RGAE,MSNet,DFAN-HAD,LREN,DeepLR,SSHAD} typically employ a straight projection by autoencoders to reconstruct the background region while neglecting anomaly targets, thereby anomalies can be detected by reconstruction error maps. Wang et al. \cite{DeepLR} incorporated a deep convolutional autoencoder with low-rank priors, simultaneously extracting discriminative attributes of anomalies while ensuring self-explainability. To improve the contrast between anomalies and backgrounds, Wang et al. \cite{autoAD} proposed an adaptive-weighted loss function, which automatically reduces the reconstruction attention of the model on potential anomaly regions. Considering the geometric structure information, Fan et al. \cite{RGAE} introduced a robust graph autoencoder detector (RGAE) to utilize supergraphs as input to preserve the geometric information of targets. Moreover, Yang et al. \cite{GEVE} argued that the relationship between pixels evolves dynamically across different scales and proposed a hierarchical guided filtering mechanism to integrate multi-scale detection results. Liu et al. \cite{MSNet} developed a multi-scale self-supervised learning network (MSNet) that employs a separate training strategy to suppress anomaly areas during the background reconstruction process. Unlike most methods that treat the background as a unified category, Cheng et al. \cite{DFAN-HAD} modeled the background as multiple pattern categories and proposed an aggregated-separation loss based on intra-class similarity and inter-class discrepancy. 

\textit{The above methods typically require heavy computations during training, resulting in an expanding computational cost in response to image scale. Different from them, our proposed method takes a detour from dense training manner to substantially ease running time while preserving precision.}

\subsection{State Space Models}
Recent advancements have demonstrated significant progress in sequence analysis by leveraging state space models \cite{hippo,S4,SSM,SSM2,SSM3}, particularly with selective structured state space models (Mamba) \cite{Mamba, Mamba2} that are distinguished by their capability to efficiently model long-range sequences with linear computational complexity. Given the exceptional proficiency of the Mamba series, it inspires the community to explore its potential within computer vision. Several representative advances \cite{Vmamba, Visionmamba} are the first to introduce Mamba into computer vision tasks by modeling long-range dependencies between image patches. Building upon this, numerous visual state space models have been developed for specific applications \cite{Videomamba,Motionmamba,Mambamil,Graphmamba,Rsmamba,MambaHSI,s2mamba}. Nevertheless, its potential in hyperspectral anomaly detection remains unexplored, which inspires us to seek a low-cost architecture for efficient hyperspectral anomaly detection. 

\section{Proposed Method}
	
\subsection{Preliminaries}
\subsubsection{Problem Formulation and Motivation}
The task of unsupervised hyperspectral anomaly detection can be described as: given a hyperspectral data cube $\mathbf{X} \in \mathbb{R}^{H\times W\times C}$, the goal is to distinguish anomaly targets and backgrounds. The prevalent approaches typically engage in an error map calculation to recognize anomaly regions, which can be generally formulated as:
\begin{equation}
    \label{eq1}
    \mathbf P=\mathbb{I}\left(\left|\mathcal D(\mathcal E(\mathbf {X}))-\mathbf{X} \right|_k >\tau\right),
\end{equation}
where $\mathcal D$ and $\mathcal E$ indicate the decoder and encoder of the detector, respectively, and $\mathbf P$ represents the detection map. $\tau$ is the detection threshold, and $\left\|*\right\|_k$ denotes the $k$-norm. 

\textit{\textbf{High Computational Cost:}} As discussed above, yielding a meticulous detection map requires a pixel-level reconstruction of HSI, which is computationally expensive. In terms of the whole image-based methods, each epoch requires calculations across $H \times W$ pixels, whereas for the patch-based methods, the computation times extends to $H \times W \times P^2$. Furthermore, existing methods leave a blank in the exploration of low-cost structures for HAD, such as state space models which are linear complexity and capable of global context learning. In contrast to previous studies, we innovatively introduce low-cost state space models into HAD and design a fast detection paradigm, significantly enhancing detection efficiency.

\textit{\textbf{Unguided Model Optimization:}} To facilitate the detection of anomaly targets, an intuitive solution is to encourage reconstruction consistency for background regions while enhancing reconstruction errors for anomaly regions. Following this criteria, existing methods typically employ data masking strategies to take detours from anomaly targets during training, thereby rendering them more difficult to reconstruct during the detection. Nevertheless, previous studies primarily adopt masking strategies on input data, which lacks a directive projection of anomalous regions. In this way, anomaly objects are unseen to the detector during training, thereby raising an inevitable reconstruction risk. Unlike existing methods, we simultaneously guide the mapping of both background and anomaly regions from the optimization perspective, enabling the model to adapt anomalies and project them into collapse space.

\subsubsection{State Space Models}
State space models (SSMs), utilizing a group of state variables to model dynamic temporal systems, have recently emerged as an efficient mechanism for analyzing sequential data, whose modeling process is formulated as a series of ordinary differential equations to perform sequence-to-sequence projections, which can be simply expressed as follows:

\begin{equation}
    \label{}
    \begin{aligned}
        & h^{\prime}(t)=\mathbf{A} h(t)+\mathbf{B} x(t) \\
        & y(t)=\mathbf{C} h(t)
    \end{aligned}
\end{equation}
where $h(t) \in \mathbb{R}^{N}$, $x(t) \in \mathbb{R}^{L}$, and $y(t) \in \mathbb{R}^{L}$ separately symbolize the latent state, input sequence, and output sequence. $N$ and $L$ are affiliated with the scale of the latent space and sequences, respectively. The core behind SSMs is to model long-range dependencies through state transition matrix $\mathbf{A} \in \mathbb{R}^{N \times N}$, control matrix $\mathbf{B} \in \mathbb{R}^{N \times L}$, and output matrix $\mathbf{C} \in \mathbb{R}^{L \times N}$. Mamba \cite{Mamba} utilizes a zero-order hold approach to convert continuous state space models into a discrete form:

\begin{equation}
    \begin{aligned}
        & \overline{\mathbf{A}}=\exp ({\Delta \mathbf{A}}) \\
        & \overline{\mathbf{B}}=({\Delta} \mathbf{A})^{-1}(\exp (\Delta \mathbf{A})-\mathbf{I}) \cdot \Delta \mathbf{B}
    \end{aligned}
\end{equation}

After discretization with the step size $\Delta$, Eq.~\ref{eq2} can be reformulated as follows: 
\begin{equation}
    \label{eq4}
    \begin{aligned}
        h_t & =\overline{\mathbf{A}} h_{t-1}+\overline{\mathbf{B}} x_t \\
        y_t & =\mathbf{C} h_t 
    \end{aligned}
\end{equation}

To facilitate the capability of modeling long-range dependencies, Mamba \cite{Mamba} outperforms previous SSMs with a selective scan mechanism that modifies the transformation matrices to data-relevant, \emph{i.e.,} $\mathbf{B} \in \mathbb{R}^{B \times L \times N}$, $\mathbf{C} \in \mathbb{R}^{B \times L \times N}$, and $\Delta \in \mathbb{R}^{B \times L \times D}$, which are associated with the $D$ dimension input data $\mathbf{X} \in \mathbb{R}^{B \times L \times D}$.

\subsection{ACMamba}
As illustrated in Figure~\ref{fig:overall}, our proposed ACMamba follows an asymmetric anomaly detection (AAD) paradigm including a region-based efficient training stage and a pixel-based meticulous detection stage. Building upon this, a lightweight autoencoder composed of regional spectral attribute learning (RSAL) modules is designed for efficient HSI reconstruction. Within the reconstruction process, we start from the perspective of optimization and introduce a consensus learning strategy (CLS), further ensuring background reconstruction and anomaly compression.

\subsubsection{\textbf{Asymmetric Anomaly Detection Paradigm}}
Considering the clumsy computational cost of the dense sampling strategy employed in existing methods, we argue that not all samples in the same homogeneous area are worthy. Conversely, acquiring the representative sample within a homogeneous region is sufficient for model building. To this end, we design an asymmetric anomaly detection paradigm to obtain both training efficiency and detection precision, involving a region-based efficient training stage and a pixel-based meticulous detection stage. 

\textbf{Region-based Efficient Training.} At the region-based efficient training stage, given a hyperspectral data cube $\mathbf{X} \in \mathbb{R}^{H \times W \times C}$, we first split it into a series of regional samples $\mathbf{\hat{X}} \in \mathbb{R}^{N_\text{r} \times C}$ using a simple linear iterative clustering algorithm \cite{slic}, each describing a homogeneous region, greatly reducing the number of training samples without impacting accuracy. Subsequently, the statistical characteristics of each region are calculated and stored in an attribute repository $\{\mathcal I^{(i)}| i \in [1, N_\text{r}]\}$, comprising mean $\mathcal I^{(i)}_{\mu}$, standard deviation $\mathcal I^{(i)}_{\sigma}$, maximum $\mathcal I^{(i)}_{\text{max}}$, and minimum $\mathcal I^{(i)}_{\text{min}}$ values, where the mean and standard deviation of region $\mathbf{\hat{X}}^{(i)}$ can be calculated as follows:
\begin{equation}
    \begin{aligned}
        & \mathcal{I}_{\mu}^{(i)}=\sum_{j=1}^{H \times W} \frac{\mathbf{X}^{(j)}\cdot\mathbb{I}(\mathbf{X}^{(j)} \in \mathbf{\hat{X}}^{(i)} )}{\sum_{j=1}^{H \times W} \mathbb{I}(\mathbf{X}^{(j)} \in \mathbf{\hat{X}}^{(i)} )}, \\ 
        & \mathcal{I}_\sigma^{(i)}=\sqrt{\frac{\sum_{j=1}^{H \times W}(\mathbf{X}^{(j)}-\mathcal{I}_\mu^{(i)})^2 \cdot \mathbb{I}(\mathbf{X}^{(j)} \in \hat{\mathbf{X}}^{(i)})}{\sum_{j=1}^{H \times W} \mathbb{I}(\mathbf{X}^{(j)} \in \hat{\mathbf{X}}^{(i)})}}.
    \end{aligned}
\end{equation}

Using this attribute repository, we can acquire representative samples $\mathbf{\tilde X} \in \mathbb{R}^{N_\text{r} \times C}$ from different areas as follows:

\begin{equation}
    \mathbf{\tilde X}^{(i)}= 
    \begin{cases} 
        \mathcal{I}_{\mu}^{(i)}+\beta \mathcal{I}_\sigma^{(i)} ,& \text{if } I^{(i)}_{\text{min}} \leq \mathcal{I}_{\mu}^{(i)}+\beta \mathcal{I}_\sigma^{(i)} \leq I^{(i)}_{\text{max}} \\
        \mathcal{I}_{\mu}^{(i)},& \text{elsewise } 
    \end{cases},
\end{equation}
where $\beta$ denotes the random factor to control the diversity of representative samples. Such samples are then fed into an autoencoder for self-supervised reconstruction: 
\begin{equation}
    \label{loss eq}
    \mathcal{L}_{\text{ACMamba}}=\frac{1}{N_\text{r}}\sum_{i=1}^{N_\text{r}}\left\|\mathbf{\tilde X}^{(i)}-\mathcal D_{\text{RSAL}}(\mathcal E_{\text{RSAL}}(\mathbf {\tilde X}^{(i)}))\right\|_k
\end{equation}
where $\mathcal D_{\text{RSAL}}$ and $\mathcal E_{\text{RSAL}}$ indicate the decoder and encoder associated with regional spectral attribute learning modules. 

\textbf{Pixel-based Meticulous Detection.} At the inference stage, region-level granularity is inadequate for achieving precise detection results, specifically for target boundaries. Consequently, we execute a pixel-based meticulous detection manner to acquire a refined detection map, which comprises two components, \emph{i.e.,} holistic and detail detection for regional instances and each pixel, respectively. The former is similar to the training stage to estimate a regional error map, where Mahalanobis distance is applied to measure non-local discrepancies. The corresponding formula is as follows:
\begin{equation}
    \mathbf{P}_{\text{holistic}}=(\mathbf{E}-\Gamma)^T\Sigma^{-1}(\mathbf{E}-\Gamma),
\end{equation}
where $\mathbf{E}$ is the regional error map between the reconstructed and original HSI. $\Gamma$ and $\Sigma$ denote its mean and standard deviation. On the other hand, we feed each pixel into the detector, by which a dense error map containing more precise results can be acquired as follows:  
\begin{equation}
    \mathbf{P}_{\text{detail}}=\left\|\mathbf{X}-\mathcal D_{\text{RSAL}}(\mathcal E_{\text{RSAL}}(\mathbf { X}))\right\|_k
\end{equation}

After that, we project holistic detection results back to $\mathbf{P}_{\text{holistic}} \in \mathbb{R}^{H \times W}$ according to the relationship between pixels and their corresponding regions. Overall, the detection map is generated by $\mathbf{P}_{\text{holistic}}\cdot\mathbf{P}_{\text{detail}}$, which effectively stands out anomalies by integrating the benefits of non-local and local cues.

\subsubsection{\textbf{Regional Spectral Attribute Learning}}
\label{RSAL}
\begin{figure}[t]
    \centering
    \begin{overpic}[width=\linewidth]{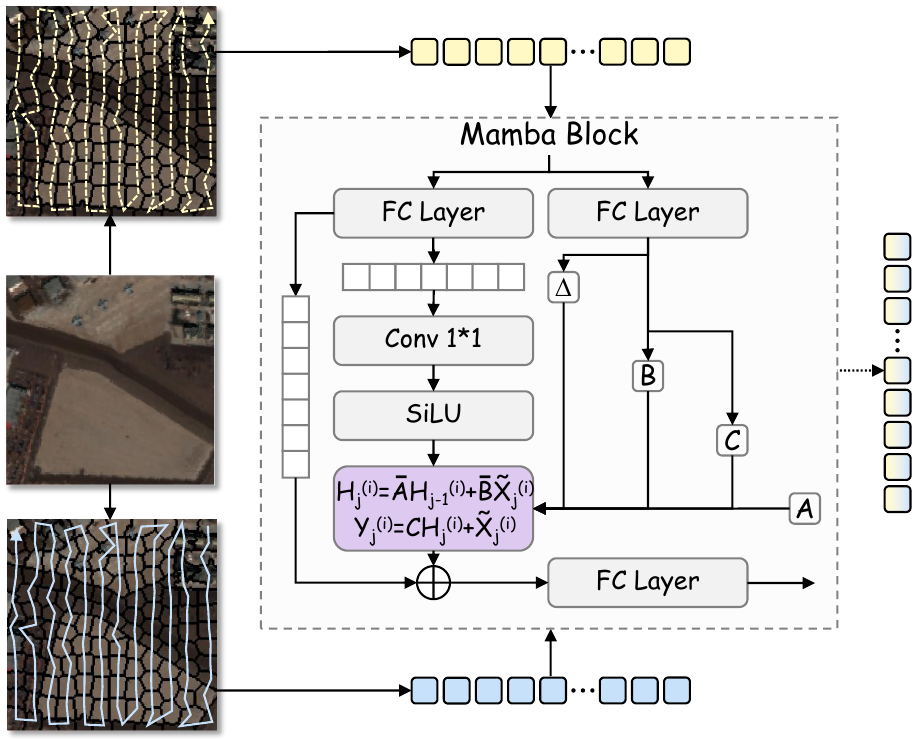}
    \put(58, 78){\scriptsize$\mathbf{\tilde X}_{0}$}    
    \put(58, -1){\scriptsize$\mathbf{\tilde X}_{1}$}       
    \put(96.5, 21){\scriptsize$\mathbf{Y}$}  
    \end{overpic}
    \caption{Illustration of the RSAL module. Its goal is capturing global contexts from regional sequences by leveraging a bi-directional scanning manner to integrate the effect from neighbors.}
    \label{fig:RSAL}
\end{figure}

To detect the anomaly targets from irregular region sequences without fixed spatial structure, we take inspiration from state space models and propose a lightweight autoencoder based on regional spectral attribute learning (RSAL) mechanisms, fully discovering global discriminative attributes by a region-by-region scanning mode. The architecture of RSAL is depicted in Figure~\ref{fig:RSAL}. Take the upper left of HSI as the starting pinpoint and unfold the sequence to obtain a series of regions $\mathbf{\tilde X} \in \mathbb{R}^{N{\text{{r}}} \times C}$. After that, we map them into two different components by a fully connected layer, which are respectively denoted with $\mathbf{x} \in \mathbb{R}^{N{\text{{r}}} \times D}$ and $\mathbf{z} \in \mathbb{R}^{N{\text{{r}}} \times D}$. The first part undergoes a convolution operation with SiLU activation and is computed by a bi-directional regional scanning mechanism to conduct an adjacent interaction:
\begin{equation}
    \begin{aligned}
        \mathbf{H}^{(i)}_{j} & =\overline{\mathbf{A}}_{\text{RSAL}} \mathbf{H}^{(i)}_{j-1}+\overline{\mathbf{B}}_{\text{RSAL}} \mathbf{\tilde X}^{(i)}_{j} \\
        \mathbf{Y}^{(i)}_{j} & =\mathbf{C}_{\text{RSAL}} \mathbf{H}^{(i)}_{j} + \mathbf{\tilde X}^{(i)}_{j} 
    \end{aligned},
\end{equation}
where $\overline{\mathbf{A}}_{\text{RSAL}}$, $\overline{\mathbf{B}}_{\text{RSAL}}$ and ${\mathbf{C}}_{\text{ RSAL}}$ represent the trainable parameters in RSAL. After scanning, we can obtain a set of output sequences:
\begin{equation}
    \mathbf{Y}=\{[\mathbf{Y}^{(j)}_{0}, \mathbf{Y}^{(j)}_{1},\ldots, \mathbf{Y}^{(j)}_{N_{\text{r}}}]\ | \mathbf{Y}^{(j)}_{0}\in \mathbb{R}^{1 \times D}, j\in \{0,1\}\},
\end{equation}

Subsequently, the output sequences derived from various scanning directions are merged based on an addition operation. Consequently, each segment of the output sequence $\mathbf{Y} \in \mathbb{R}^{N_{\text{r}} \times D}$ is capable of incorporating effects from its neighboring regions across multiple orientations. Finally, we integrate this sequence with the component $\mathbf{z}$ and feed them into a fully connected layer for dimension reduction, yielding global discriminative features.

\subsubsection{\textbf{Consensus Learning Strategy}}
Data masking strategies have recently emerged as a powerful tool for preventing the model from memorizing anomalous targets, which is crucial for boosting the detection performance. Most of existing models exclusively focuses on identity mapping for backgrounds, while neglecting the projection of anomaly targets into a collapsed space. To address this limitation, we propose a consensus learning strategy (CLS) depicted in Figure~\ref{fig:overall}, which collaboratively guides the model on background reconstruction and anomaly target suppression by dynamic gradient calibration. 

\textbf{Collaborative Reconstruction.} We extend the encoder of the detector, where two distinct encoders, \emph{i.e.,} $\mathcal E$ and $\hat{\mathcal E}$, are utilized to deal with the sequences of original regions $\mathbf{\tilde X}$ and masked regions $\mathbf M\odot \mathbf{\tilde X}$. Subsequently, a shared decoder $\mathcal D$ is employed to decode the above feature sequence into the initial HSI space, and two loss functions are performed which entirely assume the masked HSI regional sequence $\mathbf M\odot \mathbf{\tilde X}$ as the ground truth. The optimization function Eq. (\ref{loss eq}) can be rewritten as follows:
\begin{equation}
    \label{eq12}
    \begin{aligned}
        &\mathcal{L}_{\text{ACMamba}}=\mathcal{L}_{\text{ori}}+\mathcal{L}_{\text{{mask}}}\\
        &=\left\|\mathbf M\odot \mathbf{\tilde X}-\mathcal D(\mathcal E(\mathbf {\tilde X}))\right\|_k+\left\|\mathbf M\odot \mathbf{\tilde X}-\mathcal D(\hat{\mathcal E}(\mathbf M\odot \mathbf {\tilde X}))\right\|_k
    \end{aligned}
\end{equation}
where $\mathbf M \in \mathbb{R}^{N_{\text{r}}}$ is a one-hot sequence with the same length as $\mathbf{\tilde X}$. The former is tasked with projecting anomalous regions to zero while keeping the backgrounds unchanged. The latter guarantees an identity mapping across all regions. Ideally, by optimizing two components, anomaly regions should be minimized to zero, whereas the background maintains identity mapping. However, the two objective components are evidently in conflict. Thus, we take inspiration from \cite{PCGrad} and perform a gradient calibration on them. At each training batch, we randomly designate one item as the primary optimization direction and compute the angle between its gradient and the other item, \emph{i.e.,} $\theta=\arccos(\frac{\nabla{\mathcal{L}_{\text{ori}}}\cdot \nabla{\mathcal{L}_{\text{mask}}}}{\|\nabla{\mathcal{L}_{\text{ori}}}\| \|\nabla{\mathcal{L}_{\text{mask}}}\|}) $. When $\theta$ is less than $\pi/2$, their conflict can be ignored, and the overall optimization directly proceeds. On the contrary, we project the secondary gradient on the orthogonal plane of the first item to avoid conflict, such as $\nabla{\mathcal{L}_{\text{ori}}}=\nabla{\mathcal{L}_{\text{ori}}}-\frac{\nabla{\mathcal{L}_{\text{ori}}} \cdot \nabla{\mathcal{L}_{\text{mask}}}}{\left\|\nabla{\mathcal{L}_{\text{mask}}}\right\|^2} \nabla{\mathcal{L}_{\text{mask}}}$.

\textbf{Anomaly Mask Generation.} To further avoid the reconstruction of anomaly targets, we take a detour from random masking and introduce a difficulty-aware masking (DAM) approach, whose goal is to mask those regions with a high probability of belonging to anomaly targets by accumulating reconstruction errors. In particular, we build a reconstruction error list $\{\mathbf E_{1}, \mathbf E_{2},..., \mathbf E_{{N_{\text{super}}}}\}$ for collecting reconstruction error of regions at each epoch. Taking the $k$th region at the $t$th epoch as an example, its reconstruction error can be expressed:
\begin{equation}
    \mathbf{E}^{(t)}_{k}=\sum_{i=1}^{t-1}\mathbf{E}^{(i)}_{k}+\left\|\mathbf{\tilde X}-\mathcal D(\mathcal E_{\text{RSAL}}(\mathbf {\tilde X}))\right\|_k.
\end{equation}

According to this list, we randomly select $\eta\cdot N_{\text{r}}$ regions to drop, where regions with larger errors are more likely to be selected. $\eta$ is used to control the masking rate.
\begin{table*}[t]
    \newcommand{\CC}[1]{\cellcolor{gray!#1}}
    \centering
    \caption{Comparison with state-of-the-art methods on eight hyperspectral anomaly detection datasets. The best results are marked in bold.} 
    \resizebox{0.99\textwidth}{!}{\begin{tabular}{l|cccccccc|rr}
            \toprule
            Method&Urban-1 &Urban-2 & AVIRIS-1  & AVIRIS-2 & Hydice & Hyperion & Cri & HyperMap &Mean AUC~$\uparrow$&Mean Time. (s)~$\downarrow$\\
            \midrule
            \midrule
            \gr\multicolumn{11}{c}{\centering \textbf{\textit{Traditional Method}} }\\
            RX \cite{RX} & 0.9946 & 0.9692 & 0.9237 & 0.9403&0.9763 & 0.9978 & 0.9675 & 0.8240 &0.9492& 25.8952\\
            CRD \cite{CRD} & 0.9645 &  0.9029 & 0.9636 & 0.9658& 0.9398 & 0.9956 & 0.9084 & 0.7422  &0.9228&87.3286\\
            2S-GLRT \cite{2S-GLRT} & 0.9088 &  0.8530 & 0.9227 & 0.9339&0.9586 & 0.8851 &  0.8354 & 0.8355 &0.8916&1424.7725 \\
            KIFD \cite{KIFD} & 0.8533 &  0.8440 & {0.9846}& {0.9910} & 0.9961 & \textbf{0.9993} & 0.9908 & 0.5356 &0.8993&55.8886 \\
            
            \gr\multicolumn{11}{c}{\centering \textbf{\textit{Patch-based Method}} }\\
            BS3LNet \cite{BS3LNet} &  0.9526 &  0.8937 & 0.8482 & 0.9469 & 0.9538 & 0.9837 & 0.7355 & 0.5457 &0.8575&5712.3933\\
            DirectNet \cite{DirectNet} & \textbf{0.9994} &  0.9618 & 0.3154 & 0.7086& 0.6854 & 0.9767 & 0.9546 & 0.7569 &0.7948&3595.4892 \\
            GT-HAD \cite{GT-HAD} &  0.9943 & 0.9248 & 0.9475 & 0.9878& 0.9112 & 0.9903 &  0.9527 & 0.5720 &0.9101& 122.6416\\
            TAEF \cite{TAEF} &  {0.9993} & 0.9620 & 0.6455 & 0.7025& 0.6890 & 0.9825 &  0.9492 & 0.7178 &0.8310& 69.6306\\
            
            \gr\multicolumn{11}{c}{\centering \textbf{\textit{Whole Image-based Method}}} \\
            LREN \cite{LREN} &  0.9946 & 0.8708 & 0.9649 & 0.5797& 0.9536 & 0.7821 & 0.8837 & 0.6236 &0.8316& 100.0789\\
            RGAE \cite{RGAE} & 0.5574 &  0.8178 & 0.9790 & \textbf{0.9920}& 0.8396 & 0.9408 & 0.9755 & 0.4997  &0.8252&345.2650\\
            Auto-AD \cite{autoAD} &  0.9889 &  0.8493 & 0.8958 & 0.9394& 0.9970 & 0.9952 & 0.4890 & 0.5662  &0.8401&20.4618\\
            MSNet \cite{MSNet} &  0.9946 &  {0.9692} & 0.9237 & 0.9403& 0.9763 & {0.9978} & 0.9675 & 0.8241 &0.9492& 15.1609\\
            SSHAD \cite{MSNet} &  0.9720 &  0.6129 & 0.8694 & 0.8143& 0.8121 & {0.8835} & 0.9295 & 0.6679 &0.8202& 50.9944\\
            DFAN-HAD \cite{DFAN-HAD} &  0.9721 & 0.9506 & 0.9704  & 0.9857& {0.9985} & {0.9988} &0.9897 &  0.7455&0.9514&54.2796 \\
            \midrule
            
            \gr\multicolumn{11}{c}{\centering \textbf{\textit{Our Proposed Method}}} \\
            ACMamba &  {0.9967} &  \textbf{0.9770} & \textbf{0.9921} & {0.9878}&  \textbf{0.9985} & {0.9986} & \textbf{0.9986} &\textbf{0.8820} &\textbf{0.9789}& \textbf{1.5047}\\
            \bottomrule
    \end{tabular}}
    \label{table_overall}
\end{table*}

\begin{figure*}[t]
    \small
    \centering
    \begin{overpic}[width=.95\linewidth]{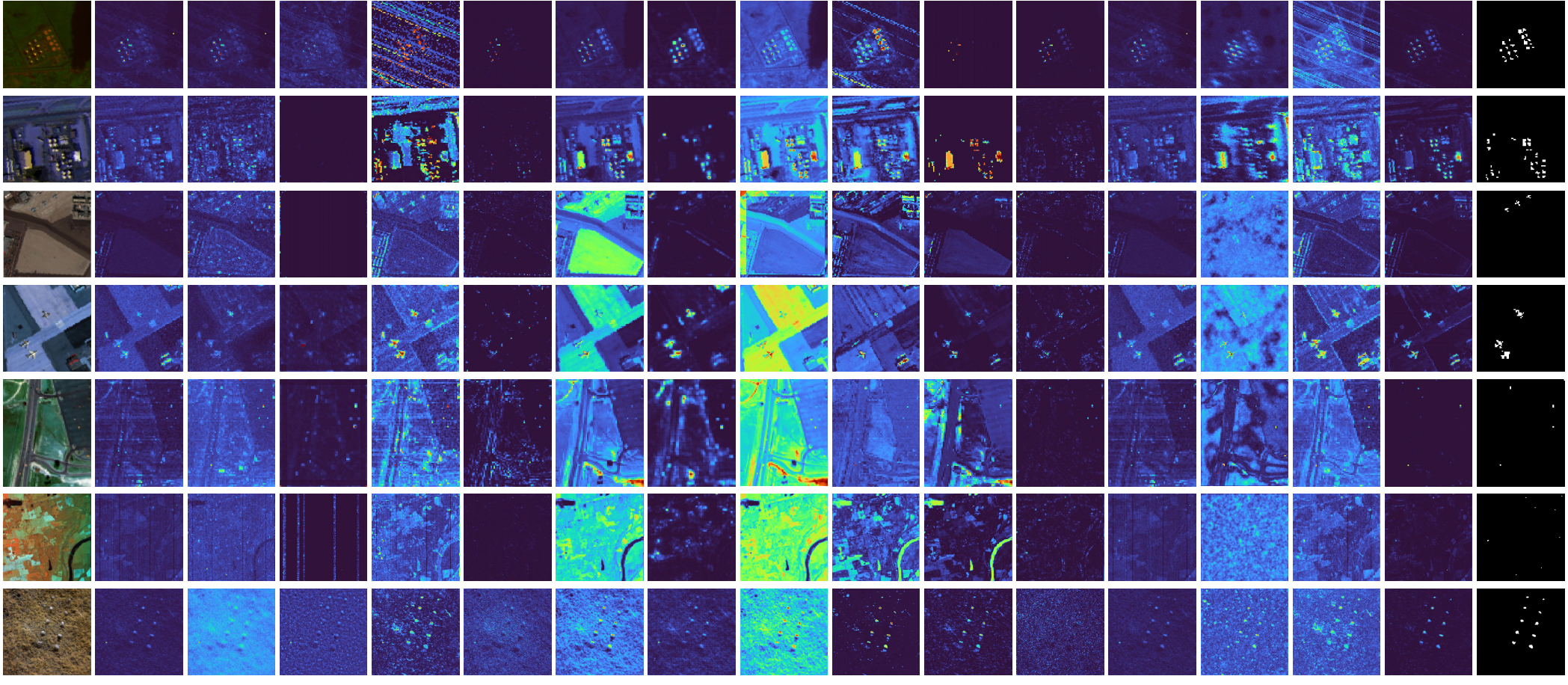} 
    \end{overpic}
    \caption{{Qualitative detection results on seven HAD datasets, from top to bottom: Urban-1, Urban-2, AVIRIS-1, AVIRIS-2, Hydice, Hyperion, and Cri. Visualizations from left to right: false-color map, RX, CRD, 2S-GLRT, KIFD, BS3LNet, DirectNet, GT-HAD, TAEF, LREN, RGAE, Auto-AD, MSNet, SSHAD, DFAN-HAD, ACMamba, and ground truth. }} 
    \label{fig:allresults}
\end{figure*}

\subsubsection{\textbf{Theoretical Analysis}}
In this section, we leverage information theory \cite{entropy-information,rethinking-autoencoder} to analyze the effectiveness of ACMamba from a theoretical perspective. The optimization objective of ACMamba aims to achieve identity mapping of backgrounds and non-identity projection of anomalies through an autoencoder framework, and anomaly detection can be performed by calculating error maps between original and reconstructed HSIs using Eq.~\ref{eq1}. Based on the hyperspectral spatial mixing property, anomalous pixels can be considered as a combination of background $\mathbf B$ and specific materials $\mathbf S$, expressed as $\mathbf A=\mathbf B+\mathbf S$, implying that the background distribution is included within the anomalous distribution. To achieve the optimal objective function, we propose Eq.~\ref{eq12}, where $\mathbf M\odot \mathbf{\tilde X}$ can be nearly treated as the background $\mathbf B$, and ${\mathbf {\tilde X}}$ represents the combination of background and anomalies $\mathbf A \cup \mathbf B$. Eq.~\ref{eq12} can be reformulated as: 
\begin{equation}
    \begin{aligned}
\mathcal{L} &=\left\|\mathbf M\odot \mathbf{\tilde X}-\mathcal D(\mathcal E(\mathbf {\tilde X}))\right\|_k+\left\|\mathbf M\odot \mathbf{\tilde X}-\mathcal D(\hat{\mathcal E}(\mathbf M\odot \mathbf {\tilde X}))\right\|_k \\
&= 2*\left\|{\mathbf B} - \mathcal D(\mathcal E({\mathbf B}))\right\|_k + \left\|{\mathbf B} - \mathcal D(\mathcal E({\mathbf A}))\right\|_k
    \end{aligned}
\end{equation}

From the information theory perspective, the mutual information between anomalies and background $\mathcal I(\mathbf A; \mathbf B) = \mathcal H(\mathbf A) - \mathcal I(\mathbf A|\mathbf B)$ leads to $\mathcal I(\mathbf A|\mathbf B) = \mathcal H(\mathbf A) - \mathcal I(\mathbf A; \mathbf B) = \mathcal H(\mathbf A) - \mathcal H(\mathbf B)$. By optimizing Eq.~\ref{eq12} to enforce $\mathcal D(\mathcal E(\mathbf B)) = \mathcal D(\mathcal E(\mathbf A))=\mathbf B$, \emph{i.e.,} $\mathcal H(\mathbf B) = \mathcal H(\mathbf A)$, we can derive $\mathcal I(\mathbf A|\mathbf B) = 0$. This demonstrates that our objective function can maximize the mutual information between anomalies and backgrounds by projecting both background and anomalous regions into background distributions, which maintains identity mapping for backgrounds while suppressing anomalies, thereby significantly enhancing detection performance.

\section{Experiments}
In this part, we first present implementation details (Section \ref{Experimentalsetup}) and then make a comprehensive comparison between ACMamba and existing state-of-the-art approaches (Section \ref{sota}). Conclusively, we perform ablation studies to analyze the effect of each component in ACMamba (Section \ref{ablation}).
\subsection{Experimental Setup}
\label{Experimentalsetup}
\subsubsection{Datasets and Evaluation}
We conduct comprehensive experiments across eight HAD datasets including Urban-1, Urban-2 \cite{ABU}, AVIRIS-1, AVIRIS-2 \cite{AVIRIS}, Hydice \cite{Hydice}, Hyperion \cite{Hyperion}, Cri \cite{Cri}, and HyperMap \cite{HyperMap} to verify the effectiveness of ACMamba. Cri and HyperMap are two large-scale datasets, leading to a time-cost challenge for HAD methods. Detailed descriptions are in the Appendix. For a fair comparison, we keep consistent with prior studies \cite{LREN,GT-HAD} that adopt the area under curve (AUC) score to evaluate the HAD performance.

\begin{table}[t]
    \newcommand{\CC}[1]{\cellcolor{gray!#1}}
    \centering
    \caption{Efficiency comparison of different methods on HyperMap.} 
    \resizebox{0.99\linewidth}{!}{\begin{tabular}{l|rr|r}
            \toprule
            Method&Train. Time (s)~$\downarrow$& Infer. Time (s)~$\downarrow$&Mean AUC~$\uparrow$\\
            \midrule
            \midrule
            \gr\multicolumn{4}{c}{\centering \textbf{\textit{Traditional Method}} }\\
            RX \cite{RX}&-&134.8473& 0.8240 \\
            KIFD \cite{KIFD}&118.1630&75.3056& 0.5362 \\
            \gr\multicolumn{4}{c}{\centering \textbf{\textit{Patch-based Method}} }\\
            DirectNet \cite{DirectNet}&2924.9061&10.8227&0.7569 \\
            TAEF \cite{TAEF} &261.7545&{15.9167 
}&0.7178 \\
            \gr\multicolumn{4}{c}{\centering \textbf{\textit{Whole Image-based Method}} }\\
            RGAE \cite{RGAE}&1865.4100&31.2700&0.4997\\
            DFAN-HAD \cite{DFAN-HAD}&339.6547&0.2672&0.7455 \\
            \midrule
            ACMamba &\textbf{4.1692}&0.6290& \textbf{0.8820}\\
            \bottomrule
    \end{tabular}}
    \label{table_time}
\end{table}

\subsubsection{Implementation Details}
We implement all the experiments on a single NVIDIA GeForce RTX 4090 GPU, employing the AdamW optimizer \cite{adamw} with a learning rate of 0.0005 to train ACMamba for 100 epochs. The number of generated regions $N_{\text{r}}$ is set to $\frac{H\times W}{\psi}$, which reduces the training cost by $\psi$ times. The trade-offs $D$, $\psi$, $\beta$, and $\eta$ are set to 256, 150, 2, and 0.01, respectively. Notably, in contrast to most existing methods that require careful hyperparameter tuning for each dataset, our ACMamba adopts unified hyperparameters to achieve satisfactory performance across different datasets.

\subsection{Comparison with State-of-the-Arts}
\label{sota}
We conduct a comprehensive comparison of ACMamba with several state-of-the-art methods. All methods are evaluated using the optimal experimental settings reported in their papers or re-implemented by their official code. 

\begin{figure}[t]
    \scriptsize
    \centering
    \includegraphics[width=\linewidth]{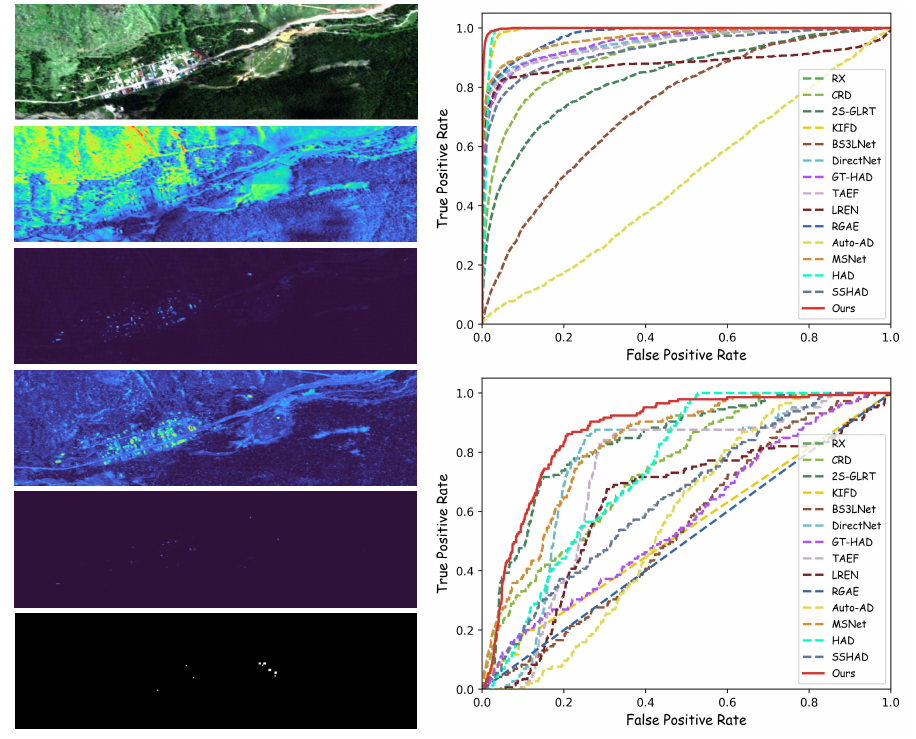}
    \caption{{Qualitative results on HyperMap. The visualizations from left to right are false-color map, TAEF, MSNet, DFAN-HAD, ACMamba, and ground truth (left). ROC curves of different methods on the Cri and HyperMap datasets (right). }}
    \label{fig:hypermap}
\end{figure}

\paragraph{Overall Performance.}In Table \ref{table_overall}, we benchmark overall performances across eight HAD datasets. To demonstrate the effectiveness of ACMamba, we compare it with efficient traditional methods and accurate deep learning methods. Table 1 reveals that our ACMamba can outperform competitors with at least 2.97\%, 6.88\%, and 2.75\% AUC gains over traditional \cite{RX}, patch-based \cite{GT-HAD}, and whole image-based methods \cite{DFAN-HAD}, demonstrating its superiority in detection precision and running speed. Compared to the state-of-the-art patch-based \cite{GT-HAD} and whole image-based \cite{MSNet,DFAN-HAD} deep learning methods, our running speed can separately achieve more than 100 and 10 times enhancements, which confirms the effectiveness and efficiency of ACMamba. Figure \ref{fig:allresults} shows detailed detection maps, which demonstrates that our ACMamba is capable of alleviating false positive detection under different scenarios. Figure \ref{fig:hypermap} explicitly demonstrate that our ACMamba can greatly suppress false-positive targets in large-scale scenarios, \emph{i.e.,} HyperMap, obtaining the most purified error map and optimal curves, proving its effectiveness.

\begin{figure*}[t]
    \scriptsize
    \centering
    \includegraphics[width=\textwidth]{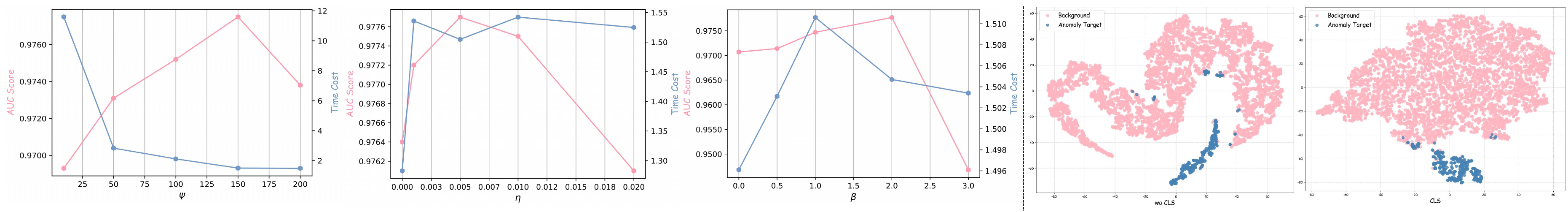}
    \caption{{Ablation on the compression ratio $\psi$ and diversity $\beta$ of representative samples, and the missing rate $\eta$ in CLS strategy (left). Feature comparison of whether adopting CLS in ACMamba on the Cri dataset (right). }}
    \label{fig.para}
\end{figure*}

\paragraph{Efficiency Comparison.}
We carry out a comparison with state-of-the-art methods in terms of detection precision and running times on the largest dataset \cite{HyperMap}, which involves 200$\times$800 pixels with 126 spectral bands. For a fair comparison, all experiments are validated on the same platform. The comparison, depicted in Table \ref{table_time}, shows that our ACMamba can significantly reduce the computational cost during training due to the proposed asymmetrical detection paradigm. In contrast, previous deep learning-based methods are hindered by heavy computation and require more than 300 seconds for deployment, confirming the efficiency of ACMamba. 

\begin{table}[t]
    \newcommand{\CC}[1]{\cellcolor{gray!#1}}
    \centering
    \caption{Ablation study of ACMamba on eight hyperspectral anomaly detection datasets. } 
    \resizebox{0.99\linewidth}{!}{\begin{tabular}{c|cccc|rr}
            \toprule
            Method&AAD&RSAL&CLS&DAM&Mean AUC~$\uparrow$&Mean Time. (s)~$\downarrow$\\
            \midrule
            \midrule
            \multirow[c]{5}{*}{ACMamba }&&&& & 0.9379& 53.3672\\
            &$\checkmark$ &&&& 0.9652 &0.9314\\
            &$\checkmark$ &$\checkmark$&&& 0.9745&\textbf{0.8628} \\
            &$\checkmark$ &$\checkmark$&$\checkmark$&& 0.9764& 1.2828\\
            \gr &$\checkmark$ &$\checkmark$  &$\checkmark$ &$\checkmark$ &\textbf{0.9789}& 1.5047 \\
            \bottomrule
    \end{tabular}}
    \label{table_ablation}
\end{table}

\begin{table}[t]
    \newcommand{\CC}[1]{\cellcolor{gray!#1}}
    \centering
    \caption{Ablation on the network architecture of ACMamba on eight hyperspectral anomaly detection datasets. } 
    \resizebox{0.99\linewidth}{!}{\begin{tabular}{c|ccc|r}
            \toprule
            Method&Depth of $\mathcal E$&Depth of $\mathcal D$&Hidden Dimension $D$&Mean AUC~$\uparrow$\\
            \midrule
            \midrule
            &1 &1&32& 0.9703 \\
            &1 &1&64& 0.9711 \\
            &1 &1&128& 0.9775 \\
            \gr ACMamba &1 &1&256& \textbf{0.9789}\\
            &1 &1  &512 &0.9752 \\
            &2 &1  &256 &{0.9732} \\
            &1 &2  &256 &{0.9672} \\
            \bottomrule
    \end{tabular}}
    \label{table_netarch}
\end{table}
\subsection{Ablation Study}
\label{ablation}
\paragraph{The effectiveness of each component in ACMamba.}
We here perform a detailed analysis of each component in ACMamba on eight HAD datasets. The baseline is built on an autoencoder architecture with multi-head attention modules utilizing the whole image as the input, where a random masking strategy is adopted on the input data. Table \ref{table_ablation} shows that introducing the AAD paradigm can improve the detection speed by 57.30 times, verifying our motivation. After integrating the RSAL module, ACMamba earns linear complexity computational capabilities and obtains a 0.93\% improvement on the AUC score. The consensus learning strategy equipped with the difficulty-aware masking strategy can further increase the detection performance, especially for large-scale scenes, \emph{i.e.,} a 2.35\% gain on HyperMap. Figure \ref{fig.para} further reveals that CLS can preserve the identity mapping of diverse backgrounds while simultaneously compressing anomalous features, confirming our claim.
\paragraph{Parameter analysis of each component in ACMamba.}
We examine a comprehensive analysis to investigate the performance of ACMamba with different hyperparameters. All the datasets are utilized to verify. Figure \ref{fig.para} studies the optimal factors $\psi$, $\beta$, and missing rate $\eta$ in data masking, which indicate the compression ratio of samples, the diversity of representative samples, and the missing rate in data masking strategy, respectively. By varying $\psi$ from 10 to 200, the time cost significantly decreases due to the compression of homogeneous regions. When $\psi$ is set to 150, ACMamba achieves the best trade-off between precision and efficiency. In contrast, $\beta$ and $\eta$ exhibit minimal effects on the time cost, whose optimal thresholds are 2.0 and 0.005, respectively. Table \ref{table_netarch} studies the optimal network architecture. When the depth of $\mathcal E$, $\mathcal D$ and $D$ are set to 1, 1 and 256, respectively, our ACMamba achieves the best results.

\section{Conclusion}
In this paper, we presented ACMamba, an efficient detection approach for hyperspectral anomaly detection. Considering not all samples within the same homogeneous area are required, ACMamba designed an asymmetrical anomaly detection paradigm by training with regional samples rather than dense pixels, vastly reducing computational costs. Under this paradigm, we introduced a regional spectral attribute learning mechanism to efficiently discover the global discriminative property of HSI regions. To substantially optimize this framework, we proposed a consensus learning strategy, which can jointly promote background reconstruction and anomaly compression. Experimental results on eight HAD datasets verify the superiority of our ACMamba.

%% file: sec/X_suppl.tex
\clearpage
\setcounter{page}{1}
\maketitlesupplementary

\section{More Dataset Details}
	
We conduct comprehensive experiments across eight HAD datasets including Urban-1, Urban-2, AVIRIS-1, AVIRIS-2, Hydice, Hyperion, Cri, and HyperMap to verify the effectiveness of ACMamba.
\begin{itemize}
    \item Urban-1 and Urban-2 \cite{ABU} are two challenging HAD datasets including roads, roofs, and buildings, whose anomaly targets are relevant to irregular buildings. The former consists of 100$\times$100 pixels with 188 spectral bands and the latter is composed of 100$\times$100 pixels with 205 spectral bands.
    \item AVIRIS-1 and AVIRIS-2 \cite{AVIRIS} are obtained by the Airborne Visible/Infrared Imaging Spectrometer (AVIRIS) sensor with the size of 150$\times$150 pixels, each pixel in AVIRIS-1 and AVIRIS-2 is composed of 186 and 204 spectral bands, respectively, whose anomalies are related to airplanes.
    \item Hydice \cite{Hydice} is obtained with the Hyperspectral Digital Image Capture Experiment (HYDICE) sensor, consisting of 80$\times$100 pixels with 175 bands, whose anomalies are regarding cars.
    \item Hyperion \cite{Hyperion} is derived from the EO-1 satellite imagery website acquired from Indiana, USA. It contains 150$\times$150 pixels with 149 spectral bands.
    \item Cri \cite{Cri} is a large-scale dataset collected by the Nuance Cri hyperspectral sensor covering an area of 400$\times$400 pixels with 46 bands, whose targets are ten rocks.
    \item HyperMap \cite{HyperMap} is a more large-scale dataset captured at the small town of Cook City, MT, USA, which involves 200$\times$800 pixels with 126 spectral bands, whose anomalies are associated with fabric panel and vehicle targets.
\end{itemize}

\section{More Detailed Related Works}
\subsection{Hyperspectral Anomaly Detection}
Hyperspectral anomaly detection emerged from the challenge of recognizing unknown targets that differ from surrounding scenes without labeled data. Traditional methods typically address it by statistics- \cite{RX,LRX,2S-GLRT,KIFD,MsRFQFT} or representation-based \cite{CRD,LRCRD,PTA,PCA-TLRSR} techniques. 

\paragraph{Statistics-based}
Reed et al. \cite{RX} introduced the Reed–Xiaoli (RX) method, a classical statistics-based method following the generalized likelihood ratio test, which assumes that backgrounds in HSI follow a multivariate Gaussian distribution and detects anomaly targets by measuring the Mahalanobis distance between pixels. Various improved methods leveraging RX as the basic pipeline have been proposed, such as local RX \cite{LRX}, weighted RX \cite{WRX}, and subspace RX \cite{SRX}.  RX-based approaches require measuring the distance between the target and backgrounds in sliding windows, however, the anomaly in practice often involves multiple pixels, resulting in backgrounds being unavailable in this case. To address this, Li et al. \cite{2S-GLRT} designed a two-step generalized likelihood ratio test method (2S-GLRT) for multiple pixel anomaly detection.  Considering anomalies are more susceptible to isolation than backgrounds in the kernel space, Li et al. \cite{KIFD} proposed a kernel isolation forest-based hyperspectral anomaly detection method (KIFD), achieving promising results by using local and global statistical information. Different from the previous methods dealing with HSI in the spatial domain, Tu et al. \cite{MsRFQFT} solved the anomaly detection task in the frequency domain by performing a Gaussian low-pass filter on the amplitude spectrum.
\begin{table*}[h]
    \newcommand{\CC}[1]{\cellcolor{gray!#1}}
    \centering
    \caption{Ablation study of ACMamba on eight hyperspectral anomaly detection datasets.} 
    \resizebox{0.99\textwidth}{!}{\begin{tabular}{cccc|cccccccc|rr}
            \toprule
            AAD&RSAL&CLS&DAM&Urban-1 &Urban-2 & AVIRIS-1  & AVIRIS-2 & Hydice & Hyperion & Cri & HyperMap&Mean AUC~$\uparrow$&Mean Time. (s)~$\downarrow$\\
            \midrule
            \midrule
            &&& & 0.9975 & 0.9344 & 0.9839 & \textbf{0.9914} & 0.9899 & 0.9681 & 0.9525 & 0.6863&0.9379& 53.3672\\
            $\checkmark$ &&&&\textbf{0.9984} & 0.9671 & 0.9891 & 0.9861 & 0.9966 & 0.9937 & 0.9868 & 0.8035 & 0.9652 &0.9314\\
            $\checkmark$ &$\checkmark$&&&0.9940 & 0.9715& 0.9911 & 0.9864 & 0.9979& 0.9981& 0.9984& 0.8585 & 0.9745&\textbf{0.8628} \\
            $\checkmark$ &$\checkmark$&$\checkmark$&&0.9946& 0.9722 & {0.9915} & 0.9862& {0.9985} & 0.9983 & {0.9983} & 0.8716 & 0.9764& 1.2828\\
            \gr $\checkmark$ &$\checkmark$  &$\checkmark$ &$\checkmark$ & {0.9967} &  \textbf{0.9770} & \textbf{0.9921} & {0.9878}&  \textbf{0.9985} & \textbf{{0.9986}} & \textbf{0.9986} &\textbf{0.8820} &\textbf{0.9789}& \textbf{1.5047} \\
            \bottomrule
    \end{tabular}}
    \label{table_allablation}
\end{table*}

\begin{figure*}[h]
    \scriptsize
    \centering
    \begin{minipage}{0.24\textwidth}
        \includegraphics[width=\linewidth]{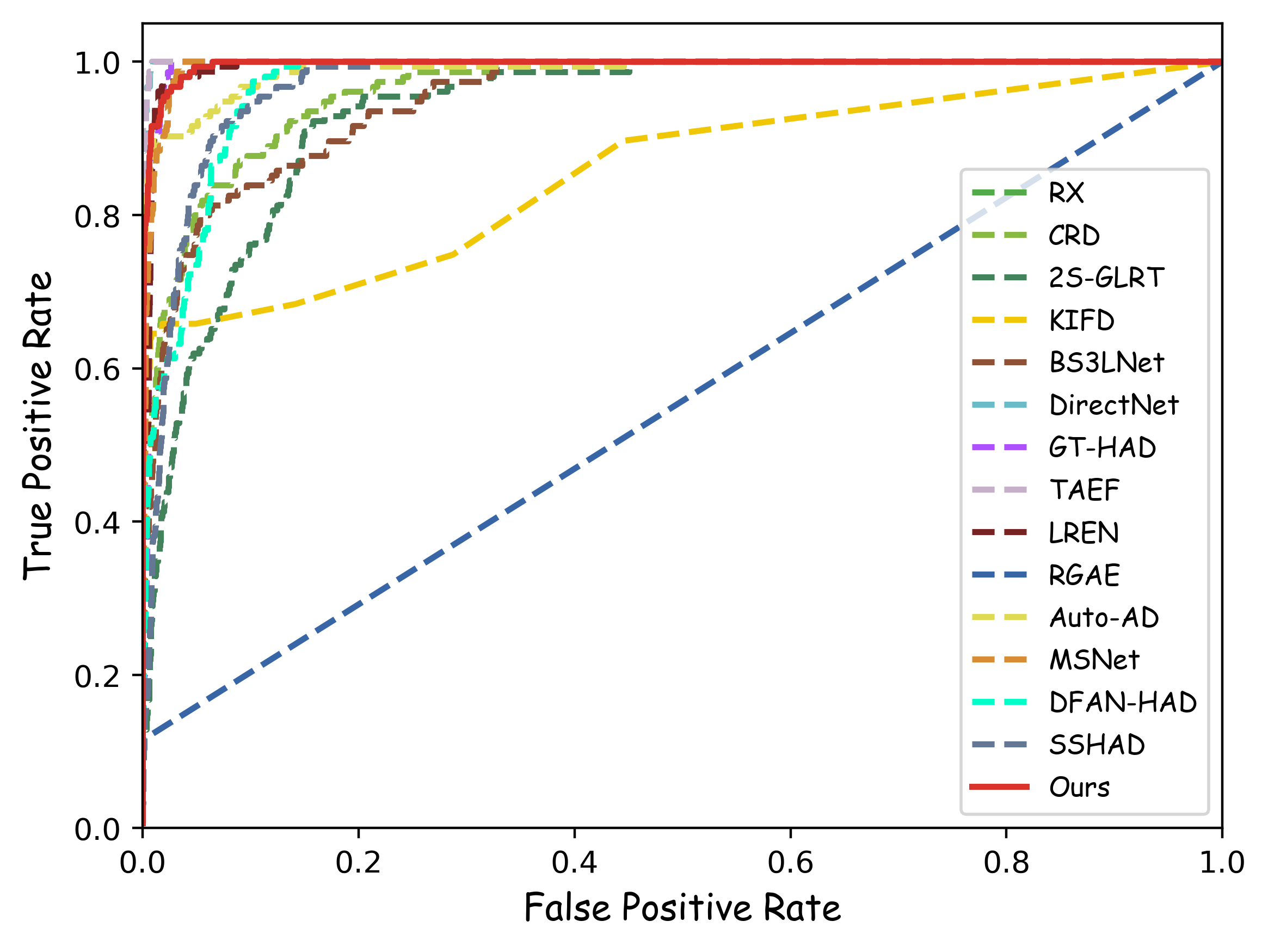}
    \end{minipage}
    \hfill
    \begin{minipage}{0.24\textwidth}
        \includegraphics[width=\textwidth]{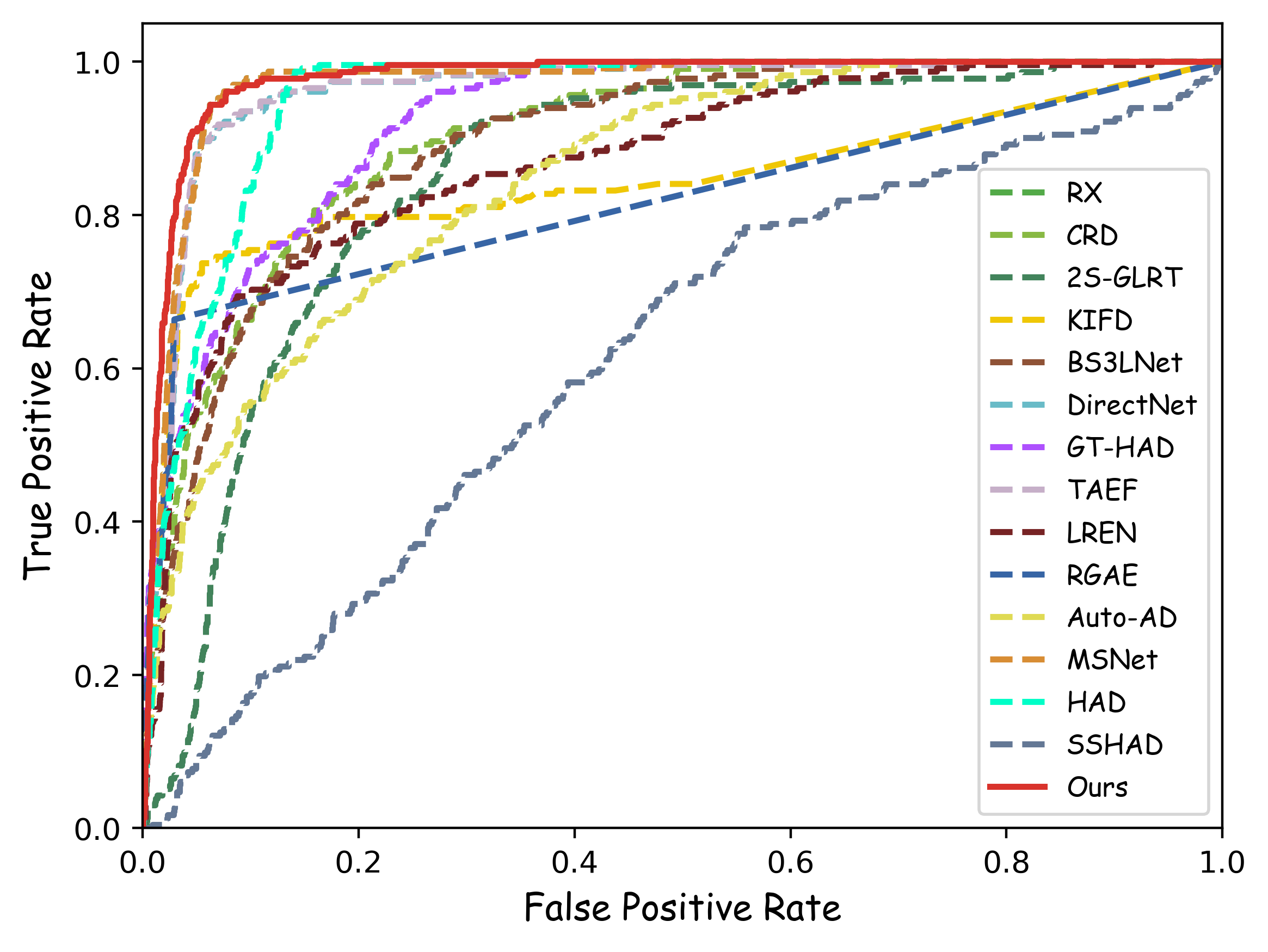}
    \end{minipage}
    \hfill
    \begin{minipage}{0.24\textwidth}
        \includegraphics[width=\textwidth]{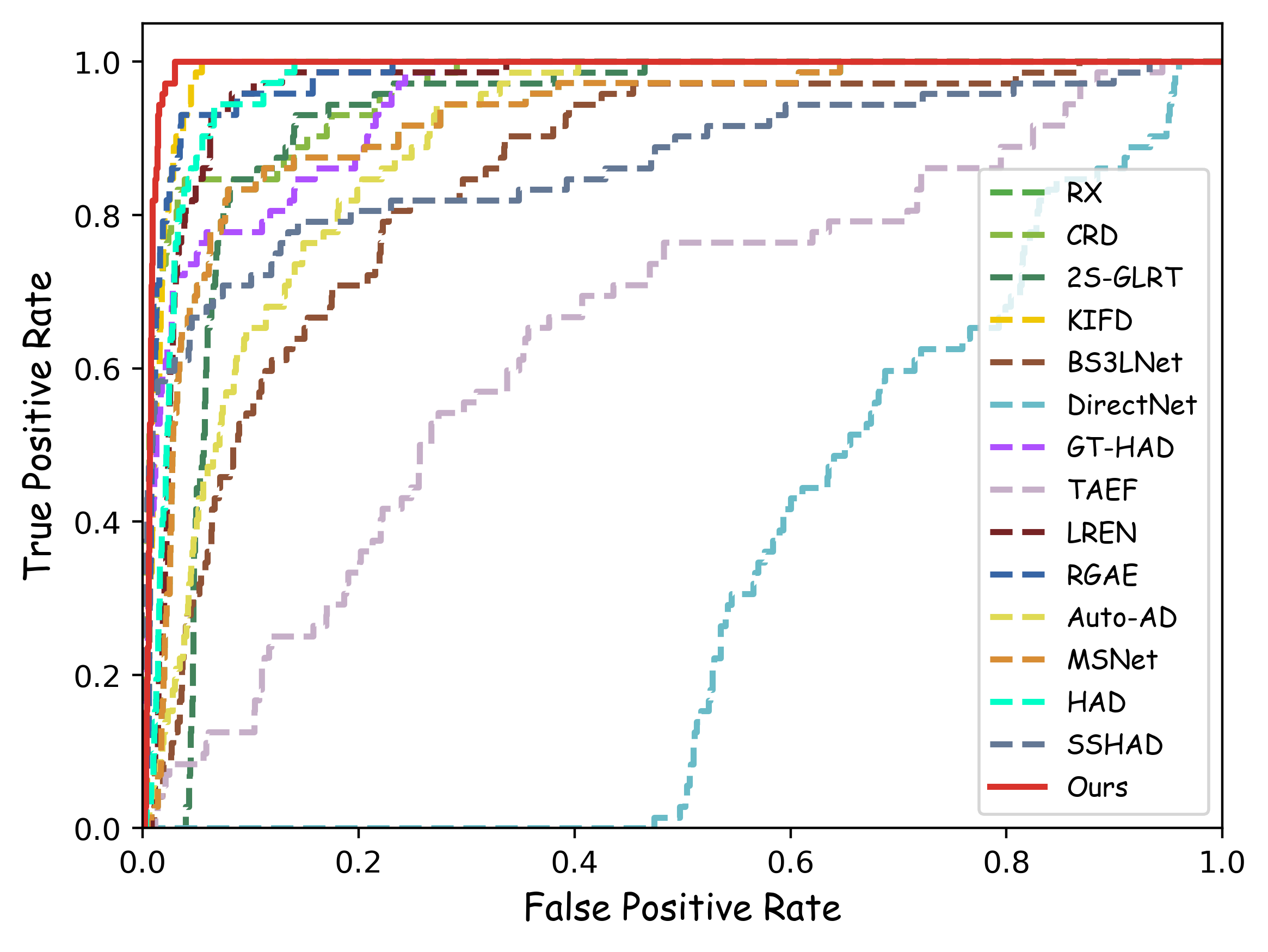}
    \end{minipage}
    \hfill
    \begin{minipage}{0.24\textwidth}
        \includegraphics[width=\textwidth]{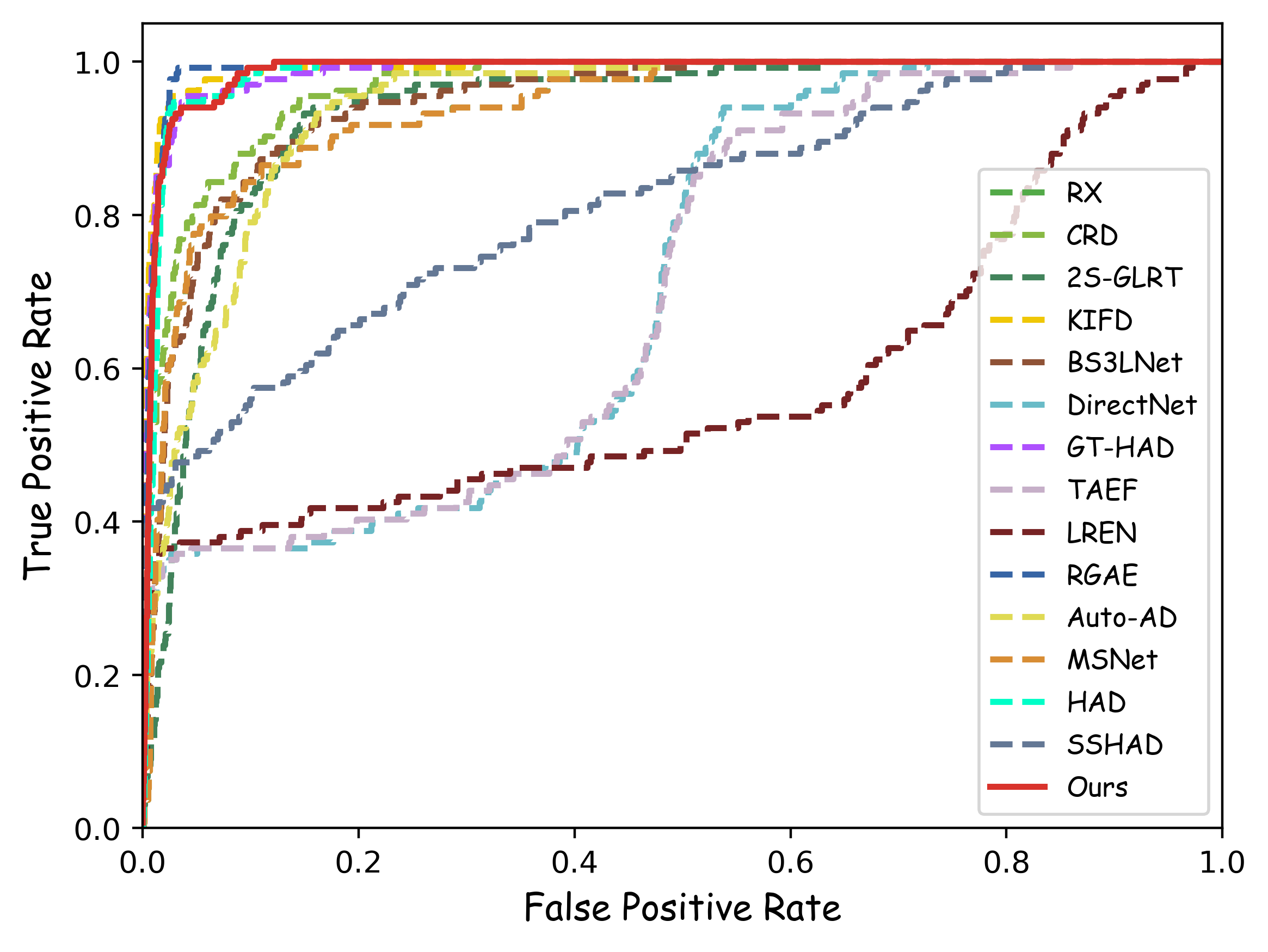}
    \end{minipage}
    \begin{minipage}{0.24\textwidth}
        \includegraphics[width=\linewidth]{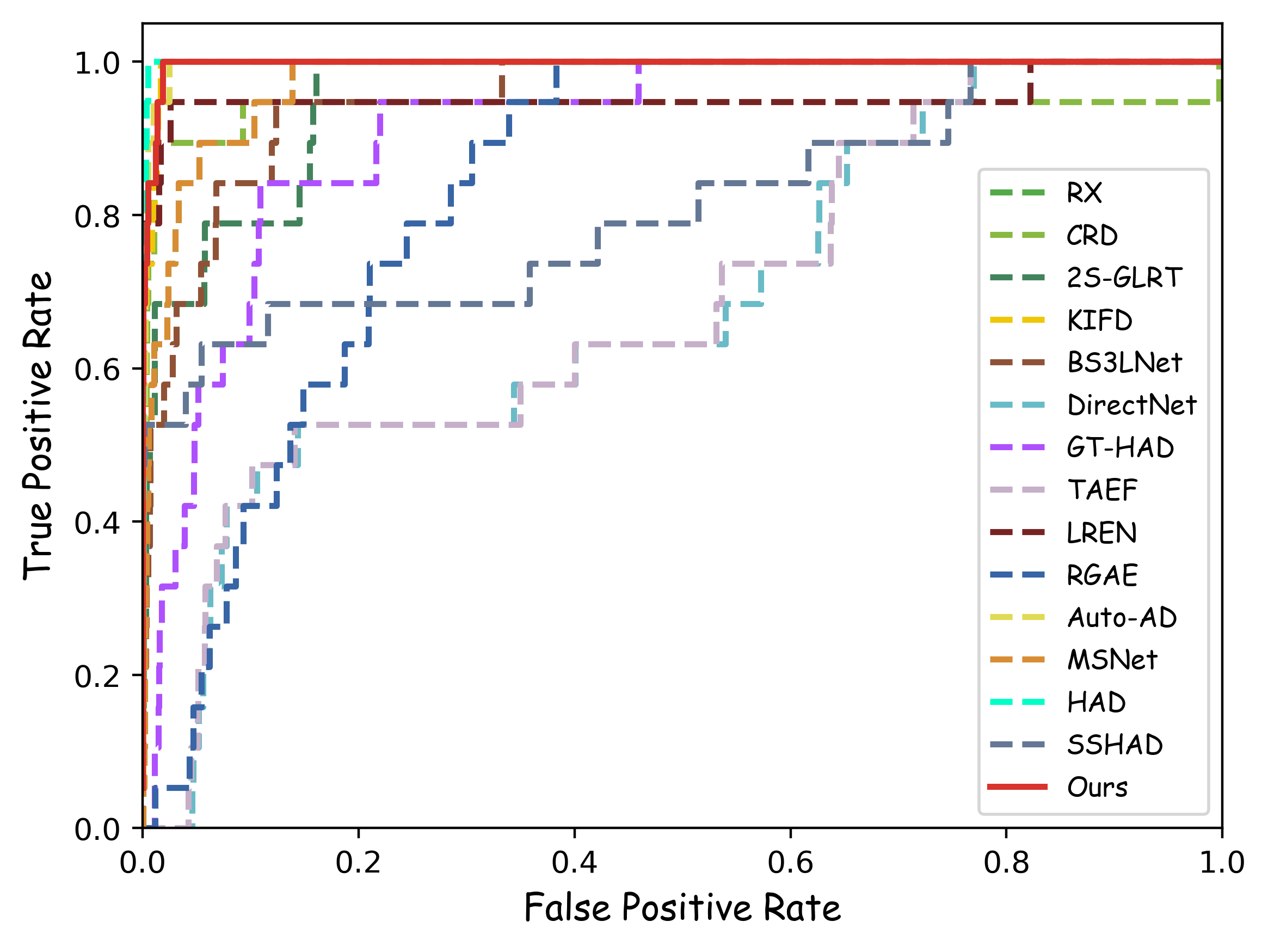}
    \end{minipage}
    \hfill
    \begin{minipage}{0.24\textwidth}
        \includegraphics[width=\textwidth]{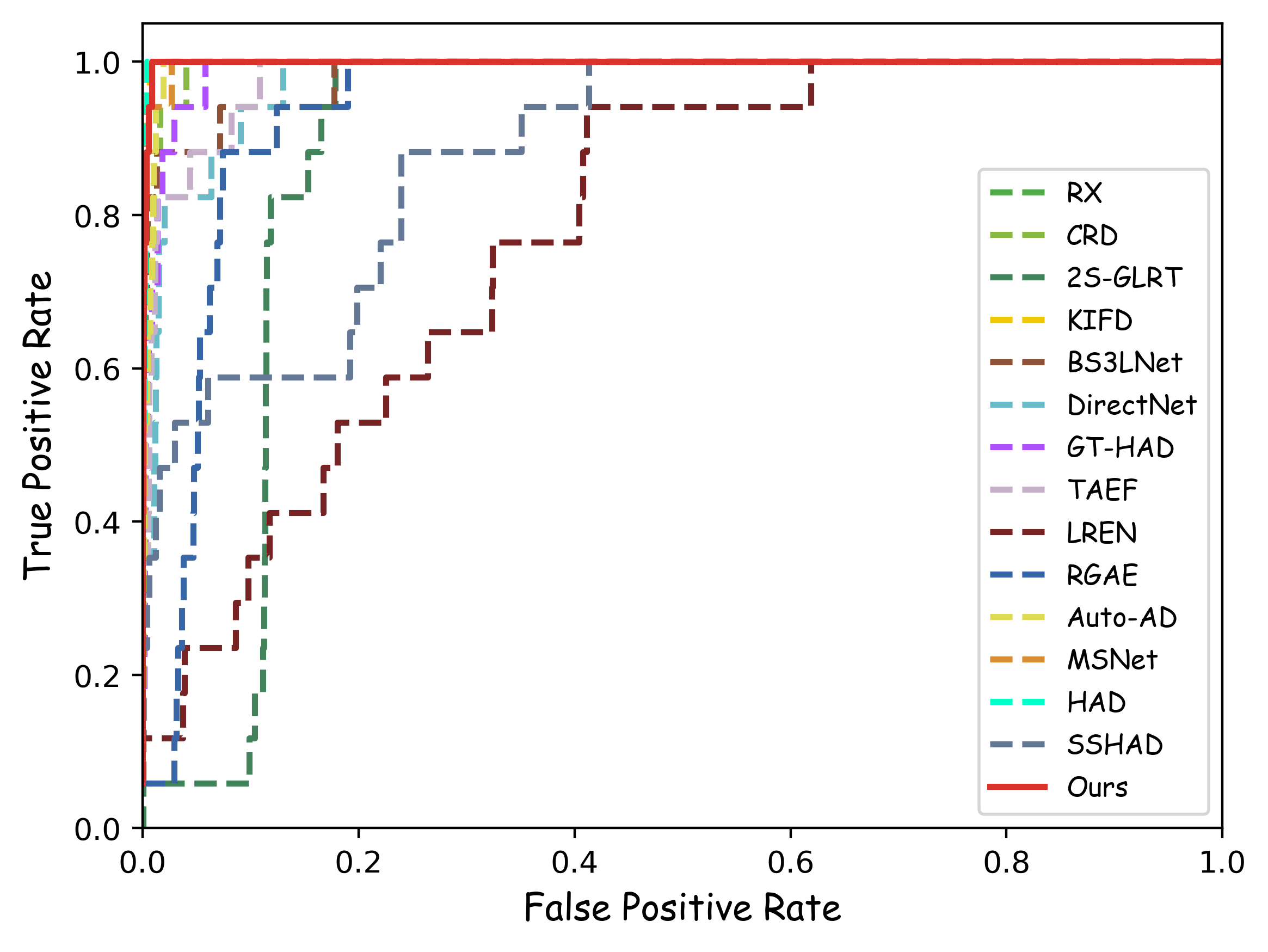}
    \end{minipage}
    \hfill
    \begin{minipage}{0.24\textwidth}
        \includegraphics[width=\textwidth]{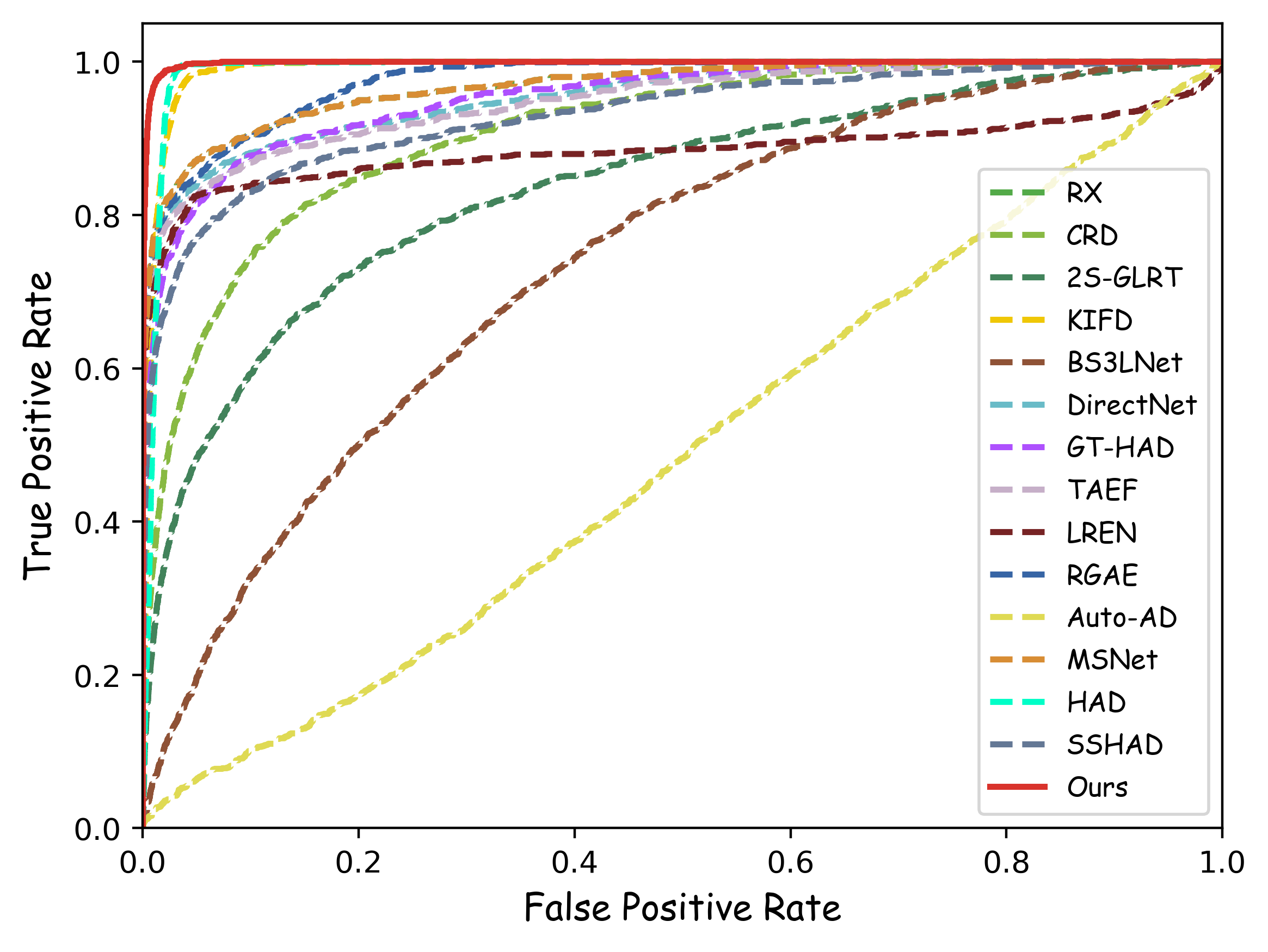}
    \end{minipage}
    \hfill
    \begin{minipage}{0.24\textwidth}
        \includegraphics[width=\textwidth]{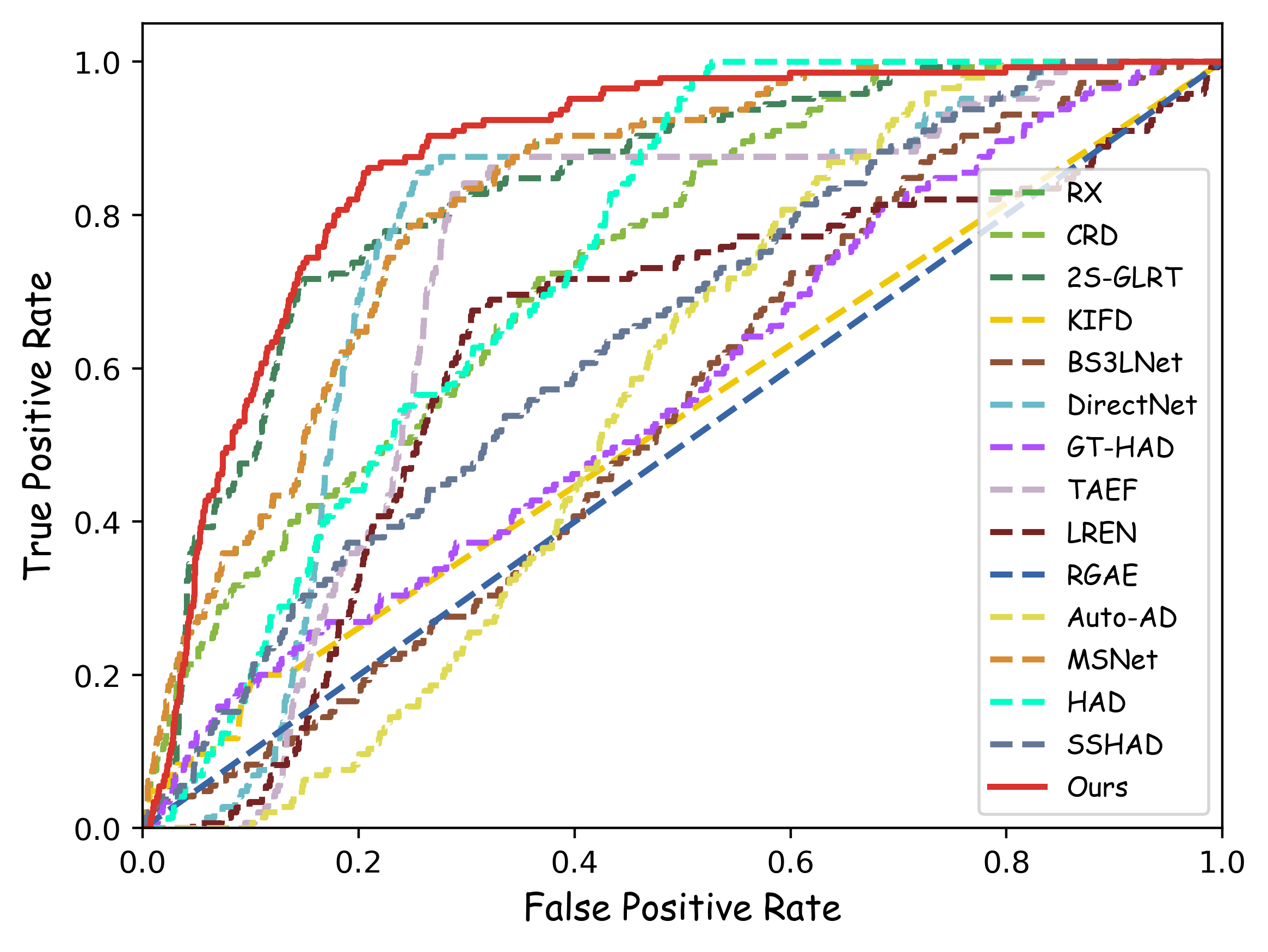}
    \end{minipage}
    \caption{{Comparison of ROC curves of different methods on eight HAD datasets. The first row is from left to right: Urban-1, Urban-2, AVIRIS-1, and AVIRIS-2. The second row is from left to right: Hydice, Hyperion, Cri, and HyperMap.}}
    \label{fig.roc}
\end{figure*}
\begin{figure*}[h]
    \small
    \centering
    \begin{overpic}[width=\linewidth]{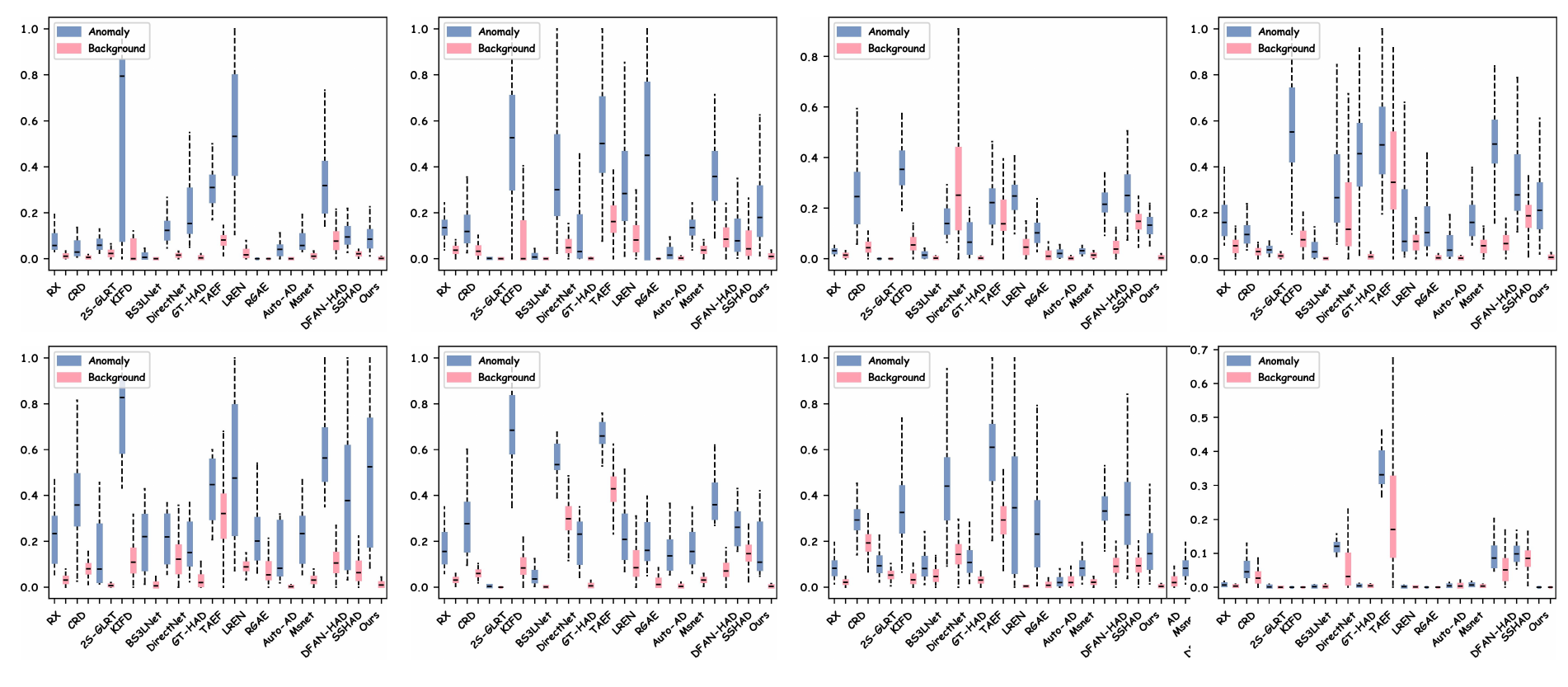} 
    \end{overpic}
    \caption{{Comparison of box plots of different methods on eight HAD datasets. The first row is from left to right: Urban-1, Urban-2, AVIRIS-1, and AVIRIS-2. The second row is from left to right: Hydice, Hyperion, Cri, and HyperMap.}} 
    \label{fig:allbox}
\end{figure*}
\begin{figure*}[h]
    \scriptsize
    \centering
    \includegraphics[width=.9\linewidth]{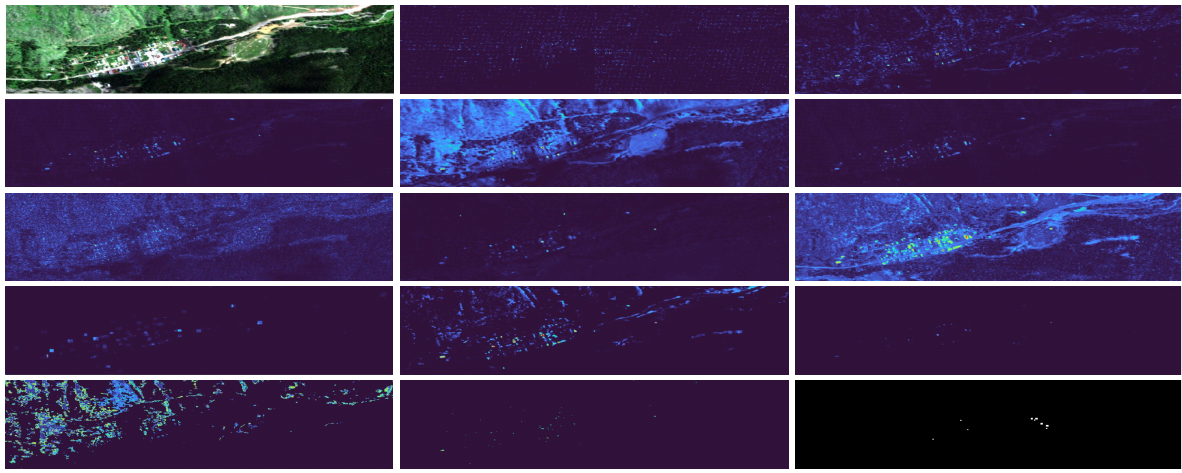}
    \caption{{Qualitative detection results on HyperMap. The first column is from top to bottom: false-color map, RX, CRD, 2S-GLRT, and KIFD. The second column is from top to bottom: BS3LNet, DirectNet, GT-HAD, LREN, and RGAE. The third column is from top to bottom: Auto-AD MSNet, DFAN-HAD, ACMamba, and ground truth. }}
    \label{fig:allhypermap}
\end{figure*}

\paragraph{Representation-based}
Representation-based methods generally assume that each pixel in backgrounds can be roughly represented by its neighborhoods in space and consider pixels that cannot be adequately described as anomaly targets. Li et al. \cite{CRD} proposed a collaborative representation-based detector (CRD), which adopts a dual-window strategy that utilizes pixels in the outer window to detect the central pixel by linear combining reconstruction. Based on CRD, Su et al. \cite{LRCRD} considered the low-rank properties of backgrounds and the sparsity of anomaly and proposed a low-rank and collaborative representation detector (LRCRD), ensuring the accuracy and stability of anomaly detection. Similarly, considering that backgrounds exhibit low-rank characteristics in homogeneous regions while anomalies show spatial sparsity, Wang et al. \cite{LELRP} proposed a locally enhanced low-rank prior (LELRP-AD) method for hyperspectral anomaly detection. In addition, Li et al. \cite{PTA} and Wang et al. \cite{PCA-TLRSR} followed low-rank and sparse-based tensor approximation to better separate backgrounds and anomalies.

\subsubsection{Deep learning Methods}
With the recent success of DL techniques in computer vision \cite{DL}, various studies introduce DL into hyperspectral image processing for anomaly detection. One of the primary distinctions among these methods is how to feed HSI data into the deep neural network for anomaly detection, roughly divided into two categories, e.g., patch-based \cite{LS3TNet,GT-HAD,DirectNet,BS3LNet,PDBSNet,AUD-Net,TAEF} and whole image-based \cite{autoAD,RGAE,MSNet,DFAN-HAD,LREN,DeepLR,SSHAD} methods.
\paragraph{Patch-based}
Patch-based methods \cite{LS3TNet, AUD-Net,BS3LNet,DirectNet,PDBSNet, GT-HAD} often split the image into numerous overlapping regions and successively feed them into the DL model for feature extraction and detection, where the information within each patch is utilized to determine whether its center pixel is an anomaly. Since most previous methods lack extraction of local spatial information, Wang et al. \cite{LS3TNet} proposed a dual-stream detector based on local spatial-spectral information aggregation, which adopts adaptive convolutions and fully connected layers to mine spatial knowledge. Considering the relationship between pixels within the HSI patch, Huyan et al. \cite{AUD-Net} proposed an anomaly detector capable of cross-image detection by relational learning. To satisfactorily reconstruct pure backgrounds, Gao et al. \cite{BS3LNet} and Wang et al. \cite{DirectNet,PDBSNet} proposed a series of hyperspectral anomaly detectors based on blind-spot networks, which reduce the awareness of the center pixel in the HSI patch, thereby preventing the reconstruction of anomaly targets. Moreover, Lian et al. \cite{GT-HAD} proposed a gated Transformer detector called GT-HAD, which enhances background features and suppresses anomaly features by using spatial-spectral content similarity, thereby performing the better background reconstructions. Wu et al. \cite{TAEF} proposed a transformer-based
autoencoder framework for nonlinear hyperspectral anomaly detection to efficiently reconstruct backgrounds, thereby improving HAD performance.

\paragraph{Whole Image-based}
Whole image-based methods \cite{autoAD,RGAE,MSNet,DFAN-HAD,LREN,DeepLR} typically employ a straight projection by autoencoders to reconstruct the background region while neglecting anomaly targets, thereby anomalies can be detected by reconstruction error maps. Wang et al. \cite{DeepLR} incorporated a deep convolutional autoencoder with low-rank priors, simultaneously extracting discriminative attributes of anomalies while ensuring self-explainability. To improve the contrast between anomalies and backgrounds, Wang et al. \cite{autoAD} proposed an adaptive-weighted loss function, which automatically reduces the reconstruction attention of the model on potential anomaly regions. Considering the geometric structure information, Fan et al. \cite{RGAE} introduced a robust graph autoencoder detector (RGAE) to utilize supergraphs as input to preserve the geometric information of targets. Moreover, Yang et al. \cite{GEVE} argued that the relationship between pixels evolves dynamically across different scales and proposed a hierarchical guided filtering mechanism to integrate multi-scale detection results. Liu et al. \cite{MSNet} developed a multi-scale self-supervised learning network (MSNet) that employs a separate training strategy to suppress anomaly areas during the background reconstruction process. Unlike most methods that treat the background as a unified category, Cheng et al. \cite{DFAN-HAD} modeled the background as multiple pattern categories and proposed an aggregated-separation loss based on intra-class similarity and inter-class discrepancy. Liu et al. \cite{SSHAD} designed a spatial–spectral adaptive dual-domain learning framework, which attempt to leverage selected state space modules to extract discriminative features of HSI.

\section{More Experimental Results}
We here provide more detailed experimental results across eight HAD datasets to verify the effectiveness of ACMamba. We frist make a comperhensive analysis of ablation studies for each HAD dataset, as depicted in Table \ref{table_allablation}. Then, we provide more detailed ROC curves and box plots of all the compared methods on all benchmarks to demonstrate our effectiveness. Finally, we provide more visualizations on the HyperMap dataset, which exhibits our ACMamba can still achieve the best results.

%% file: main.bbl
\begin{thebibliography}{61}
\providecommand{\natexlab}[1]{#1}
\providecommand{\url}[1]{\texttt{#1}}
\expandafter\ifx\csname urlstyle\endcsname\relax
  \providecommand{\doi}[1]{doi: #1}\else
  \providecommand{\doi}{doi: \begingroup \urlstyle{rm}\Url}\fi

\bibitem[Achanta et~al.(2012)Achanta, Shaji, Smith, Lucchi, Fua, and Süsstrunk]{slic}
Radhakrishna Achanta, Appu Shaji, Kevin Smith, Aurelien Lucchi, Pascal Fua, and Sabine Süsstrunk.
\newblock Slic superpixels compared to state-of-the-art superpixel methods.
\newblock \emph{IEEE Transactions on Pattern Analysis and Machine Intelligence}, 34\penalty0 (11):\penalty0 2274--2282, 2012.

\bibitem[Behrouz and Hashemi(2024)]{Graphmamba}
Ali Behrouz and Farnoosh Hashemi.
\newblock Graph mamba: Towards learning on graphs with state space models.
\newblock In \emph{Proceedings of the 30th ACM SIGKDD Conference on Knowledge Discovery and Data Mining}, pages 119--130, 2024.

\bibitem[Bodrito et~al.(2021)Bodrito, Zouaoui, Chanussot, and Mairal]{NIPS-HSI}
Th{\'e}o Bodrito, Alexandre Zouaoui, Jocelyn Chanussot, and Julien Mairal.
\newblock A trainable spectral-spatial sparse coding model for hyperspectral image restoration.
\newblock \emph{Advances in Neural Information Processing Systems}, 34:\penalty0 5430--5442, 2021.

\bibitem[Cai et~al.(2024)Cai, Chen, and Cheng]{rethinking-autoencoder}
Yu Cai, Hao Chen, and Kwang-Ting Cheng.
\newblock Rethinking autoencoders for medical anomaly detection from a theoretical perspective.
\newblock In \emph{International Conference on Medical Image Computing and Computer-Assisted Intervention}, pages 544--554. Springer, 2024.

\bibitem[Chen et~al.(2024)Chen, Chen, Liu, Li, Zou, and Shi]{Rsmamba}
Keyan Chen, Bowen Chen, Chenyang Liu, Wenyuan Li, Zhengxia Zou, and Zhenwei Shi.
\newblock Rsmamba: Remote sensing image classification with state space model.
\newblock \emph{IEEE Geoscience and Remote Sensing Letters}, 21:\penalty0 1--5, 2024.

\bibitem[Cheng et~al.(2024)Cheng, Huo, Lin, Dong, Zhao, Zhang, and Wang]{DFAN-HAD}
Xi Cheng, Yu Huo, Sheng Lin, Youqiang Dong, Shaobo Zhao, Min Zhang, and Hai Wang.
\newblock Deep feature aggregation network for hyperspectral anomaly detection.
\newblock \emph{IEEE Transactions on Instrumentation and Measurement}, 73:\penalty0 1--16, 2024.

\bibitem[Dao and Gu(2024)]{Mamba2}
Tri Dao and Albert Gu.
\newblock Transformers are ssms: Generalized models and efficient algorithms through structured state space duality.
\newblock In \emph{Forty-first International Conference on Machine Learning}, 2024.

\bibitem[Dosovitskiy et~al.(2020)Dosovitskiy, Beyer, Kolesnikov, Weissenborn, Zhai, Unterthiner, Dehghani, Minderer, Heigold, Gelly, et~al.]{ViT}
Alexey Dosovitskiy, Lucas Beyer, Alexander Kolesnikov, Dirk Weissenborn, Xiaohua Zhai, Thomas Unterthiner, Mostafa Dehghani, Matthias Minderer, Georg Heigold, Sylvain Gelly, et~al.
\newblock An image is worth 16x16 words: Transformers for image recognition at scale.
\newblock \emph{arXiv preprint arXiv:2010.11929}, 2020.

\bibitem[Du and Zhang(2014)]{Hyperion}
Bo Du and Liangpei Zhang.
\newblock A discriminative metric learning based anomaly detection method.
\newblock \emph{IEEE Transactions on Geoscience and Remote Sensing}, 52\penalty0 (11):\penalty0 6844--6857, 2014.

\bibitem[Fan et~al.(2021)Fan, Ma, Mei, Fan, Huang, and Ma]{RGAE}
Ganghui Fan, Yong Ma, Xiaoguang Mei, Fan Fan, Jun Huang, and Jiayi Ma.
\newblock Hyperspectral anomaly detection with robust graph autoencoders.
\newblock \emph{IEEE Transactions on Geoscience and Remote Sensing}, 60:\penalty0 1--14, 2021.

\bibitem[Gabri{\'e} et~al.(2018)Gabri{\'e}, Manoel, Luneau, Macris, Krzakala, Zdeborov{\'a}, et~al.]{entropy-information}
Marylou Gabri{\'e}, Andre Manoel, Cl{\'e}ment Luneau, Nicolas Macris, Florent Krzakala, Lenka Zdeborov{\'a}, et~al.
\newblock Entropy and mutual information in models of deep neural networks.
\newblock \emph{Advances in neural information processing systems}, 31, 2018.

\bibitem[Gao et~al.(2023)Gao, Wang, Zhuang, Sun, Huang, and Plaza]{BS3LNet}
Lianru Gao, Degang Wang, Lina Zhuang, Xu Sun, Min Huang, and Antonio Plaza.
\newblock Bs3lnet: A new blind-spot self-supervised learning network for hyperspectral anomaly detection.
\newblock \emph{IEEE Transactions on Geoscience and Remote Sensing}, 61:\penalty0 1--18, 2023.

\bibitem[Green et~al.(1998)Green, Eastwood, Sarture, Chrien, Aronsson, Chippendale, Faust, Pavri, Chovit, Solis, Olah, and Williams]{AVIRIS}
Robert~O Green, Michael~L Eastwood, Charles~M Sarture, Thomas~G Chrien, Mikael Aronsson, Bruce~J Chippendale, Jessica~A Faust, Betina~E Pavri, Christopher~J Chovit, Manuel Solis, Martin~R Olah, and Orlesa Williams.
\newblock Imaging spectroscopy and the airborne visible/infrared imaging spectrometer (aviris).
\newblock \emph{Remote Sensing of Environment}, 65\penalty0 (3):\penalty0 227--248, 1998.

\bibitem[Gu and Dao(2023)]{Mamba}
Albert Gu and Tri Dao.
\newblock Mamba: Linear-time sequence modeling with selective state spaces.
\newblock \emph{arXiv preprint arXiv:2312.00752}, 2023.

\bibitem[Gu et~al.(2020)Gu, Dao, Ermon, Rudra, and R{\'e}]{hippo}
Albert Gu, Tri Dao, Stefano Ermon, Atri Rudra, and Christopher R{\'e}.
\newblock Hippo: Recurrent memory with optimal polynomial projections.
\newblock \emph{Advances in neural information processing systems}, 33:\penalty0 1474--1487, 2020.

\bibitem[Gu et~al.(2021{\natexlab{a}})Gu, Goel, and Re]{S4}
Albert Gu, Karan Goel, and Christopher Re.
\newblock Efficiently modeling long sequences with structured state spaces.
\newblock In \emph{International Conference on Learning Representations}, 2021{\natexlab{a}}.

\bibitem[Gu et~al.(2021{\natexlab{b}})Gu, Johnson, Goel, Saab, Dao, Rudra, and R{\'e}]{SSM}
Albert Gu, Isys Johnson, Karan Goel, Khaled Saab, Tri Dao, Atri Rudra, and Christopher R{\'e}.
\newblock Combining recurrent, convolutional, and continuous-time models with linear state space layers.
\newblock \emph{Advances in neural information processing systems}, 34:\penalty0 572--585, 2021{\natexlab{b}}.

\bibitem[Guo et~al.(2014)Guo, Zhang, Ran, Gao, Li, and Plaza]{WRX}
Qiandong Guo, Bing Zhang, Qiong Ran, Lianru Gao, Jun Li, and Antonio Plaza.
\newblock Weighted-rxd and linear filter-based rxd: Improving background statistics estimation for anomaly detection in hyperspectral imagery.
\newblock \emph{IEEE Journal of Selected Topics in Applied Earth Observations and Remote Sensing}, 7\penalty0 (6):\penalty0 2351--2366, 2014.

\bibitem[Gupta et~al.(2022)Gupta, Gu, and Berant]{SSM3}
Ankit Gupta, Albert Gu, and Jonathan Berant.
\newblock Diagonal state spaces are as effective as structured state spaces.
\newblock \emph{Advances in Neural Information Processing Systems}, 35:\penalty0 22982--22994, 2022.

\bibitem[Huyan et~al.(2022)Huyan, Zhang, Quan, Chanussot, and Jiao]{AUD-Net}
Ning Huyan, Xiangrong Zhang, Dou Quan, Jocelyn Chanussot, and Licheng Jiao.
\newblock Aud-net: A unified deep detector for multiple hyperspectral image anomaly detection via relation and few-shot learning.
\newblock \emph{IEEE Transactions on Neural Networks and Learning Systems}, 35\penalty0 (5):\penalty0 6835--6849, 2022.

\bibitem[Jiang et~al.(2021)Jiang, Xie, Lei, Jiang, and Li]{LREN}
Kai Jiang, Weiying Xie, Jie Lei, Tao Jiang, and Yunsong Li.
\newblock Lren: Low-rank embedded network for sample-free hyperspectral anomaly detection.
\newblock In \emph{Proceedings of the AAAI Conference on Artificial Intelligence}, pages 4139--4146, 2021.

\bibitem[Kang et~al.(2017)Kang, Zhang, Li, Li, Li, and Benediktsson]{ABU}
Xudong Kang, Xiangping Zhang, Shutao Li, Kenli Li, Jun Li, and J{\'o}n~Atli Benediktsson.
\newblock Hyperspectral anomaly detection with attribute and edge-preserving filters.
\newblock \emph{IEEE Transactions on Geoscience and Remote Sensing}, 55\penalty0 (10):\penalty0 5600--5611, 2017.

\bibitem[Kwon et~al.(2003)Kwon, Der, and Nasrabadi]{LRX}
Heesung Kwon, Sandor~Z Der, and Nasser~M Nasrabadi.
\newblock Adaptive anomaly detection using subspace separation for hyperspectral imagery.
\newblock \emph{Optical Engineering}, 42\penalty0 (11):\penalty0 3342--3351, 2003.

\bibitem[LeCun et~al.(2015)LeCun, Bengio, and Hinton]{DL}
Yann LeCun, Yoshua Bengio, and Geoffrey Hinton.
\newblock Deep learning.
\newblock \emph{nature}, 521\penalty0 (7553):\penalty0 436--444, 2015.

\bibitem[Li et~al.(2015)Li, Zhang, Zhang, and Ma]{Hydice}
Jiayi Li, Hongyan Zhang, Liangpei Zhang, and Li Ma.
\newblock Hyperspectral anomaly detection by the use of background joint sparse representation.
\newblock \emph{IEEE Journal of Selected Topics in Applied Earth Observations and Remote Sensing}, 8\penalty0 (6):\penalty0 2523--2533, 2015.

\bibitem[Li et~al.(2025)Li, Li, Wang, He, Wang, Wang, and Qiao]{Videomamba}
Kunchang Li, Xinhao Li, Yi Wang, Yinan He, Yali Wang, Limin Wang, and Yu Qiao.
\newblock Videomamba: State space model for efficient video understanding.
\newblock In \emph{European Conference on Computer Vision}, pages 237--255. Springer, 2025.

\bibitem[Li et~al.(2020)Li, Li, Qu, Zhao, Tao, and Du]{PTA}
Lu Li, Wei Li, Ying Qu, Chunhui Zhao, Ran Tao, and Qian Du.
\newblock Prior-based tensor approximation for anomaly detection in hyperspectral imagery.
\newblock \emph{IEEE Transactions on Neural Networks and Learning Systems}, 33\penalty0 (3):\penalty0 1037--1050, 2020.

\bibitem[Li et~al.(2023)Li, Fu, and Zhang]{AAAI-HSI}
Miaoyu Li, Ying Fu, and Yulun Zhang.
\newblock Spatial-spectral transformer for hyperspectral image denoising.
\newblock In \emph{Proceedings of the AAAI Conference on Artificial Intelligence}, pages 1368--1376, 2023.

\bibitem[Li et~al.(2019)Li, Zhang, Duan, and Kang]{KIFD}
Shutao Li, Kunzhong Zhang, Puhong Duan, and Xudong Kang.
\newblock Hyperspectral anomaly detection with kernel isolation forest.
\newblock \emph{IEEE Transactions on Geoscience and Remote Sensing}, 58\penalty0 (1):\penalty0 319--329, 2019.

\bibitem[Li and Du(2014)]{CRD}
Wei Li and Qian Du.
\newblock Collaborative representation for hyperspectral anomaly detection.
\newblock \emph{IEEE Transactions on geoscience and remote sensing}, 53\penalty0 (3):\penalty0 1463--1474, 2014.

\bibitem[Li et~al.(2024)Li, Luo, Zhang, Wang, and Du]{MambaHSI}
Yapeng Li, Yong Luo, Lefei Zhang, Zengmao Wang, and Bo Du.
\newblock Mambahsi: Spatial-spectral mamba for hyperspectral image classification.
\newblock \emph{IEEE Transactions on Geoscience and Remote Sensing}, 2024.

\bibitem[Lian et~al.(2024)Lian, Wang, Sun, and Huang]{GT-HAD}
Jie Lian, Lizhi Wang, He Sun, and Hua Huang.
\newblock Gt-had: Gated transformer for hyperspectral anomaly detection.
\newblock \emph{IEEE Transactions on Neural Networks and Learning Systems}, 2024.

\bibitem[Liu et~al.(2024{\natexlab{a}})Liu, Su, Shen, and Zhou]{MSNet}
Haijun Liu, Xi Su, Xiangfei Shen, and Xichuan Zhou.
\newblock Msnet: Self-supervised multi-scale network with enhanced separation training for hyperspectral anomaly detection.
\newblock \emph{IEEE Transactions on Geoscience and Remote Sensing}, 2024{\natexlab{a}}.

\bibitem[Liu et~al.(2021)Liu, Hou, Li, Tao, Orlando, and Li]{2S-GLRT}
Jun Liu, Zengfu Hou, Wei Li, Ran Tao, Danilo Orlando, and Hongbin Li.
\newblock Multipixel anomaly detection with unknown patterns for hyperspectral imagery.
\newblock \emph{IEEE Transactions on Neural Networks and Learning Systems}, 33\penalty0 (10):\penalty0 5557--5567, 2021.

\bibitem[Liu et~al.(2025)Liu, Peng, Chang, Wen, and Zhu]{SSHAD}
Sitian Liu, Lintao Peng, Xuyang Chang, Guanghui Wen, and Chunli Zhu.
\newblock Adaptive dual-domain learning for hyperspectral anomaly detection with state space models.
\newblock \emph{IEEE Transactions on Geoscience and Remote Sensing}, 2025.

\bibitem[Liu et~al.(2024{\natexlab{b}})Liu, Tian, Zhao, Yu, Xie, Wang, Ye, and Liu]{Vmamba}
Yue Liu, Yunjie Tian, Yuzhong Zhao, Hongtian Yu, Lingxi Xie, Yaowei Wang, Qixiang Ye, and Yunfan Liu.
\newblock Vmamba: Visual state space model.
\newblock \emph{Advances in Neural Information Processing Systems}, 2024{\natexlab{b}}.

\bibitem[Loshchilov and Hutter(2018)]{adamw}
Ilya Loshchilov and Frank Hutter.
\newblock Decoupled weight decay regularization.
\newblock In \emph{International Conference on Learning Representations}, 2018.

\bibitem[Peng et~al.(2023)Peng, Alcaide, Anthony, Albalak, Arcadinho, Biderman, Cao, Cheng, Chung, Derczynski, et~al.]{RWKV}
Bo Peng, Eric Alcaide, Quentin~Gregory Anthony, Alon Albalak, Samuel Arcadinho, Stella Biderman, Huanqi Cao, Xin Cheng, Michael~Nguyen Chung, Leon Derczynski, et~al.
\newblock Rwkv: Reinventing rnns for the transformer era.
\newblock In \emph{Empirical Methods in Natural Language Processing}, 2023.

\bibitem[Reed and Yu(1990)]{RX}
Irving~S Reed and Xiaoli Yu.
\newblock Adaptive multiple-band cfar detection of an optical pattern with unknown spectral distribution.
\newblock \emph{IEEE transactions on acoustics, speech, and signal processing}, 38\penalty0 (10):\penalty0 1760--1770, 1990.

\bibitem[Smith et~al.(2022)Smith, Warrington, and Linderman]{SSM2}
Jimmy~TH Smith, Andrew Warrington, and Scott Linderman.
\newblock Simplified state space layers for sequence modeling.
\newblock In \emph{The Eleventh International Conference on Learning Representations}, 2022.

\bibitem[Snyder et~al.(2008)Snyder, Kerekes, Fairweather, Crabtree, Shive, and Hager]{HyperMap}
D. Snyder, J. Kerekes, I. Fairweather, R. Crabtree, J. Shive, and S. Hager.
\newblock Development of a web-based application to evaluate target finding algorithms.
\newblock In \emph{IGARSS 2008 - 2008 IEEE International Geoscience and Remote Sensing Symposium}, pages II--915--II--918, 2008.

\bibitem[Su et~al.(2020)Su, Wu, Zhu, and Du]{LRCRD}
Hongjun Su, Zhaoyue Wu, A-Xing Zhu, and Qian Du.
\newblock Low rank and collaborative representation for hyperspectral anomaly detection via robust dictionary construction.
\newblock \emph{ISPRS Journal of Photogrammetry and Remote Sensing}, 169:\penalty0 195--211, 2020.

\bibitem[Su et~al.(2021)Su, Wu, Zhang, and Du]{HAD-review}
Hongjun Su, Zhaoyue Wu, Huihui Zhang, and Qian Du.
\newblock Hyperspectral anomaly detection: A survey.
\newblock \emph{IEEE Geoscience and Remote Sensing Magazine}, 10\penalty0 (1):\penalty0 64--90, 2021.

\bibitem[Tu et~al.(2023)Tu, Yang, He, Li, and Plaza]{MsRFQFT}
Bing Tu, Xianchang Yang, Wei He, Jun Li, and Antonio Plaza.
\newblock Hyperspectral anomaly detection using reconstruction fusion of quaternion frequency domain analysis.
\newblock \emph{IEEE Transactions on Neural Networks and Learning Systems}, 2023.

\bibitem[Vaswani et~al.(2017)Vaswani, Shazeer, Parmar, Uszkoreit, Jones, Gomez, Kaiser, and Polosukhin]{transformer}
Ashish Vaswani, Noam Shazeer, Niki Parmar, Jakob Uszkoreit, Llion Jones, Aidan~N Gomez, {\L}ukasz Kaiser, and Illia Polosukhin.
\newblock Attention is all you need.
\newblock \emph{Advances in neural information processing systems}, 30, 2017.

\bibitem[Wang et~al.(2023)Wang, Zhuang, Gao, Sun, Huang, and Plaza]{PDBSNet}
Degang Wang, Lina Zhuang, Lianru Gao, Xu Sun, Min Huang, and Antonio~J Plaza.
\newblock Pdbsnet: Pixel-shuffle downsampling blind-spot reconstruction network for hyperspectral anomaly detection.
\newblock \emph{IEEE Transactions on Geoscience and Remote Sensing}, 61:\penalty0 1--14, 2023.

\bibitem[Wang et~al.(2024)Wang, Zhuang, Gao, Sun, Zhao, and Plaza]{DirectNet}
Degang Wang, Lina Zhuang, Lianru Gao, Xu Sun, Xiaobin Zhao, and Antonio Plaza.
\newblock Sliding dual-window-inspired reconstruction network for hyperspectral anomaly detection.
\newblock \emph{IEEE Transactions on Geoscience and Remote Sensing}, 2024.

\bibitem[Wang et~al.(2025)Wang, Zhang, Peng, Zhang, and Jiao]{s2mamba}
Guanchun Wang, Xiangrong Zhang, Zelin Peng, Tianyang Zhang, and Licheng Jiao.
\newblock S2mamba: A spatial-spectral state space model for hyperspectral image classification.
\newblock \emph{IEEE Transactions on Geoscience and Remote Sensing}, pages 1--1, 2025.

\bibitem[Wang et~al.(2022{\natexlab{a}})Wang, Wang, Hong, Roy, and Chanussot]{PCA-TLRSR}
Minghua Wang, Qiang Wang, Danfeng Hong, Swalpa~Kumar Roy, and Jocelyn Chanussot.
\newblock Learning tensor low-rank representation for hyperspectral anomaly detection.
\newblock \emph{IEEE Transactions on Cybernetics}, 53\penalty0 (1):\penalty0 679--691, 2022{\natexlab{a}}.

\bibitem[Wang et~al.(2020)Wang, Wang, Zhong, and Zhang]{LELRP}
Shaoyu Wang, Xinyu Wang, Yanfei Zhong, and Liangpei Zhang.
\newblock Hyperspectral anomaly detection via locally enhanced low-rank prior.
\newblock \emph{IEEE Transactions on Geoscience and Remote Sensing}, 58\penalty0 (10):\penalty0 6995--7009, 2020.

\bibitem[Wang et~al.(2021)Wang, Wang, Zhang, and Zhong]{autoAD}
Shaoyu Wang, Xinyu Wang, Liangpei Zhang, and Yanfei Zhong.
\newblock Auto-ad: Autonomous hyperspectral anomaly detection network based on fully convolutional autoencoder.
\newblock \emph{IEEE Transactions on Geoscience and Remote Sensing}, 60:\penalty0 1--14, 2021.

\bibitem[Wang et~al.(2022{\natexlab{b}})Wang, Wang, Zhang, and Zhong]{DeepLR}
Shaoyu Wang, Xinyu Wang, Liangpei Zhang, and Yanfei Zhong.
\newblock Deep low-rank prior for hyperspectral anomaly detection.
\newblock \emph{IEEE Transactions on Geoscience and Remote Sensing}, 60:\penalty0 1--17, 2022{\natexlab{b}}.

\bibitem[Wang et~al.(2022{\natexlab{c}})Wang, Wang, and Wang]{LS3TNet}
Xiaoyi Wang, Liguo Wang, and Qunming Wang.
\newblock Local spatial-spectral information-integrated semisupervised two-stream network for hyperspectral anomaly detection.
\newblock \emph{IEEE Transactions on Geoscience and Remote Sensing}, 60:\penalty0 1--15, 2022{\natexlab{c}}.

\bibitem[Wu and Wang(2024)]{TAEF}
Ziyu Wu and Bin Wang.
\newblock Transformer-based autoencoder framework for nonlinear hyperspectral anomaly detection.
\newblock \emph{IEEE Transactions on Geoscience and Remote Sensing}, 62:\penalty0 1--15, 2024.

\bibitem[Yang et~al.(2024)Yang, Wang, and Chen]{Mambamil}
Shu Yang, Yihui Wang, and Hao Chen.
\newblock Mambamil: Enhancing long sequence modeling with sequence reordering in computational pathology.
\newblock In \emph{International Conference on Medical Image Computing and Computer-Assisted Intervention}, pages 296--306. Springer, 2024.

\bibitem[Yang et~al.(2023)Yang, Tu, Li, Li, and Plaza]{GEVE}
Xianchang Yang, Bing Tu, Qianming Li, Jun Li, and Antonio Plaza.
\newblock Graph evolution-based vertex extraction for hyperspectral anomaly detection.
\newblock \emph{IEEE Transactions on Neural Networks and Learning Systems}, 2023.

\bibitem[Yu et~al.(2020)Yu, Kumar, Gupta, Levine, Hausman, and Finn]{PCGrad}
Tianhe Yu, Saurabh Kumar, Abhishek Gupta, Sergey Levine, Karol Hausman, and Chelsea Finn.
\newblock Gradient surgery for multi-task learning.
\newblock \emph{Advances in Neural Information Processing Systems}, 33:\penalty0 5824--5836, 2020.

\bibitem[Zhang et~al.(2016)Zhang, Du, Zhang, and Wang]{Cri}
Yuxiang Zhang, Bo Du, Liangpei Zhang, and Shugen Wang.
\newblock A low-rank and sparse matrix decomposition-based mahalanobis distance method for hyperspectral anomaly detection.
\newblock \emph{IEEE Transactions on Geoscience and Remote Sensing}, 54\penalty0 (3):\penalty0 1376--1389, 2016.

\bibitem[Zhang et~al.(2025)Zhang, Liu, Reid, Hartley, Zhuang, and Tang]{Motionmamba}
Zeyu Zhang, Akide Liu, Ian Reid, Richard Hartley, Bohan Zhuang, and Hao Tang.
\newblock Motion mamba: Efficient and long sequence motion generation.
\newblock In \emph{European Conference on Computer Vision}, pages 265--282. Springer, 2025.

\bibitem[Zhao et~al.(2014)Zhao, Du, and Zhang]{SRX}
Rui Zhao, Bo Du, and Liangpei Zhang.
\newblock A robust nonlinear hyperspectral anomaly detection approach.
\newblock \emph{IEEE Journal of Selected Topics in Applied Earth Observations and Remote Sensing}, 7\penalty0 (4):\penalty0 1227--1234, 2014.

\bibitem[Zhu et~al.(2024)Zhu, Liao, Zhang, Wang, Liu, and Wang]{Visionmamba}
Lianghui Zhu, Bencheng Liao, Qian Zhang, Xinlong Wang, Wenyu Liu, and Xinggang Wang.
\newblock Vision mamba: Efficient visual representation learning with bidirectional state space model.
\newblock In \emph{Forty-first International Conference on Machine Learning}, 2024.

\end{thebibliography}
